\documentclass[10pt,journal,compsoc]{IEEEtran}
\ifCLASSOPTIONcompsoc
  \usepackage[nocompress]{cite}
\else
  \usepackage{cite}
\fi
\ifCLASSINFOpdf
\else
\fi

\usepackage[utf8]{inputenc} 
\usepackage[T1]{fontenc}    
\usepackage{hyperref}       
\usepackage{url}            
\usepackage{booktabs}       
\usepackage{amsfonts}       
\usepackage{nicefrac}       
\usepackage{microtype}      
\usepackage{graphicx}
\usepackage{multirow}
\newcommand{\vect}[1]{{\bf #1}}

\newcommand{\al}{{\mathbf{a}}}
\newcommand{\sh}{{\mathbf{s}}}
\newcommand{\co}{{\bf{c}}}
\newcommand{\alp}{{\hat{\mathbf{a}}}}
\newcommand{\shp}{{\hat{\mathbf{s}}}}
\newcommand{\cop}{{\hat{\bf{c}}}}
\newcommand{\shm}{{\mathit{s}}}

\newcommand{\im}{{\mathit{I}}}

\newcommand{\iat}{{\mathit{a}}}
\newcommand{\ist}{{\mathit{s}}}

\newcommand{\ipt}{{\mathsf p}}
\newcommand{\rb}{{\mathsf R}}
\newcommand{\tb}{{\mathsf T}}
\newcommand{\rtu}{{\mathsf r}}
\newcommand{\ttu}{{\mathsf t}}
\newcommand{\ea} {{\em et al.}}

\newcommand{\todo}[1]{{{\bf TODO:} {\em {#1}}}}
\newcommand{\ignore}[1]{}
\begin{document}
%
\title{Intrinsic Image Decomposition Using Paradigms}

%

\author{%
  D.A. Forsyth and Jason J. Rock}
\markboth{Journal of \LaTeX\ Class Files,~Vol.~14, No.~8, August~2015}%
{Shell \MakeLowercase{\textit{et al.}}: Bare Demo of IEEEtran.cls for Computer Society Journals}
%



\IEEEtitleabstractindextext{%
\begin{abstract}
Intrinsic image decomposition is the classical task of mapping image
to albedo.  The WHDR dataset allows methods to be evaluated 
by comparing predictions to human judgements (``lighter'', ``same as'',
``darker'').   The best modern intrinsic image methods learn a map from image to albedo using
rendered models and human judgements. This is convenient for practical methods, but
cannot explain how a visual agent without geometric, surface and
illumination models and a renderer could learn to recover intrinsic
images. 

This paper describes a method that learns intrinsic image decomposition
without seeing WHDR annotations, rendered data, or ground truth data.
The method relies on paradigms - fake albedos and fake shading fields
- together with a novel smoothing procedure that ensures good behavior
at short scales on real images.  Long scale error is controlled by
averaging.   Our method achieves WHDR scores competitive with those of
strong recent methods allowed to see training WHDR annotations, 
rendered data, and ground truth data.    Because our method is
unsupervised, we can compute estimates of the test/train variance of
WHDR scores; these are quite large, and it is unsafe to rely small 
differences in reported WHDR.  
\end{abstract}

}

\maketitle

\IEEEdisplaynontitleabstractindextext

%
\IEEEpeerreviewmaketitle

\IEEEraisesectionheading{\section{Introduction}\label{sec:introduction}}

%
%
%
%

Many computer vision problems can be thought of as regressing some spatial fields (for example, normal; depth; shading;
albedo; a processed image; a denoised image; etc) against an image.    An extremely powerful strategy for solving such
problems is to collect a large set of representative tuples and use them to train a convolutional neural network.   This supervised
strategy runs into difficulties when it is hard to obtain data.   We
use predicting albedo and shading from an image (otherwise, intrinsic images) as a model problem, because
the problem is well understood and well studied and because there are strong established criteria for evaluation.
For intrinsic images, the supervised framework is unsatisfying for several reasons.  First, it is hard to obtain reliable
data.  Second, until recently quite unsophisticated unsupervised methods were competitive with supervised methods,
suggesting that more sophisticated unsupervised methods worth studying.  Finally, supervised methods cannot explain
how a visual agent might learn to produce intrinsic images without ever having seen an intrinsic image -- visual animals are not
provided with true or rendered albedo data at conception.

Intrinsic images have several important properties.  An intrinsic image decomposition should {\bf explain} the image --
pixel values should be accurately predicted by albedo and shading.
Intrinsic images have a {\bf local character} -- one can tell whether a
moderately sized image patch is an albedo (resp. shading) patch without reference to the rest of the albedo
(resp. shading) field.  Similarly, the mapping from an image to an intrinsic image has a local character -- for a large
enough image patch, the intrinsic images recovered from the patch should be the same as those recovered from the whole
image. Furthermore, the mapping from an image to an intrinsic image is
{\bf equivariant} under image translation and orthonormal transformations
-- for any two images of the same scene, the albedo (resp. shading) reported for overlapping regions should be the same.
Similarly, the mapping from an image to an intrinsic image is somewhat scale equivariant  -- a
moderate scaling of an image up or down should result in intrinsic images that are similarly scaled up or down. 

Our method exploits these properties.  We train a network to decompose fixed size {\em tiles} ($128\times 128$ in this
paper) to albedo and shading estimates.  We have no ground truth, but the local character of the problem means we can
train a network using synthetic albedo (resp. shading) fields that need be accurate models only locally.    
If we pass a real tile through this network, its reported albedo (resp. shading) fields should ``look like'' the
synthetic fields locally, too, and we achieve this with an adversarial loss.  Furthermore, the albedo and shading fields
should explain the image, so we penalize the residual with a loss. To ensure that we report a translation equivariant
estimate, we cover a real image with randomly offset, overlapping tiles, compute albedo (resp. shading) fields for the
tiles, then average, so the albedo at a given location is an average over all tiles covering that location.  We apply
this procedure over several scales and average the result to obtain scale equivariance.  We compute a scale and
translation equivariant estimate for a discrete set of rotations and reflections, and average those to estimate a rotation
equivariant result.   Finally, a simple pointwise procedure ensures that the residual is small.

\section{Related work}


Originally, an intrinsic image decomposition reduced an image to
intrinsic components (properties of surfaces like albedo, specular
albedo, roughness) and  extrinsic components (like irradiance or
shading)~\cite{BarTen78}.  We adopt current usage, which implies a decomposition into
albedo and shading.   

In 1959, Edwin Land described intrinsic image procedures that estimated albedo at an image location by accumulating
comparisons~\cite{nn8146,nn8147}. Land modelled images as shaded Mondrians --- albedos were modelled as piecewise
constant patches of color and shading as a smooth field.   One then concludes that albedo displays large (but
no small) image gradients, and that shading has small (but no large) gradients.  This assumption results in a class of
methods that: compute image gradients; recover albedo gradients from the image gradients (typically, by testing gradient
magnitude); then recovering the albedo from the albedo gradients (typically, by a form of integration).    This assumption or variants underly numerous
algorithms for recovering albedo, which typically differ by how the albedo gradients are identified and by how
albedo is recovered from putative gradients (which are not directly integrable)
\cite{mccann1976quantitative,horn1974determining,horn1973lightness,Blake,BrelstaffBlake,kimmel2003variational,elad03,weiss2001deriving,levin2004separating,Levin:2007ho,zhao2012closed, 
  Tappen:2005hy,brainard1997bayesian,cc42423,brainard:1986an,Bousseau:2009ec}. 
Some variants use tuned prior models \cite{Chang:2014ti,Gehler:2011vc,shen2011intrinsic,Shen:2011gf}, or user intervention \cite{Levin:2007ho,Bousseau:2009ec}. 
Coupling to shape models appears to significantly improve shading and reflectance estimation \cite{Barron:2013uc}.  
\cite{Sheng20} use an unsupervised energy based framework, with two distinct shading components.

This paper shares essential features with Retinex-like models.  First, this work is unsupervised (or minimally supervised, if one uses data to choose a gradient magnitude threshold).
Second, this work assumes that the key questions in recovering intrinsic images is deciding whether local phenomena are
due to albedo or to shading (as in gradient thresholding), and then assembling a global estimate from those decisions
(as in integration).  In contrast to Retinex-like models, rather than use abstract models of albedo and shading to
motivate formulations, samples from these models are used to train a decomposition procedure. 

\begin{figure*}
\includegraphics[width=\textwidth]{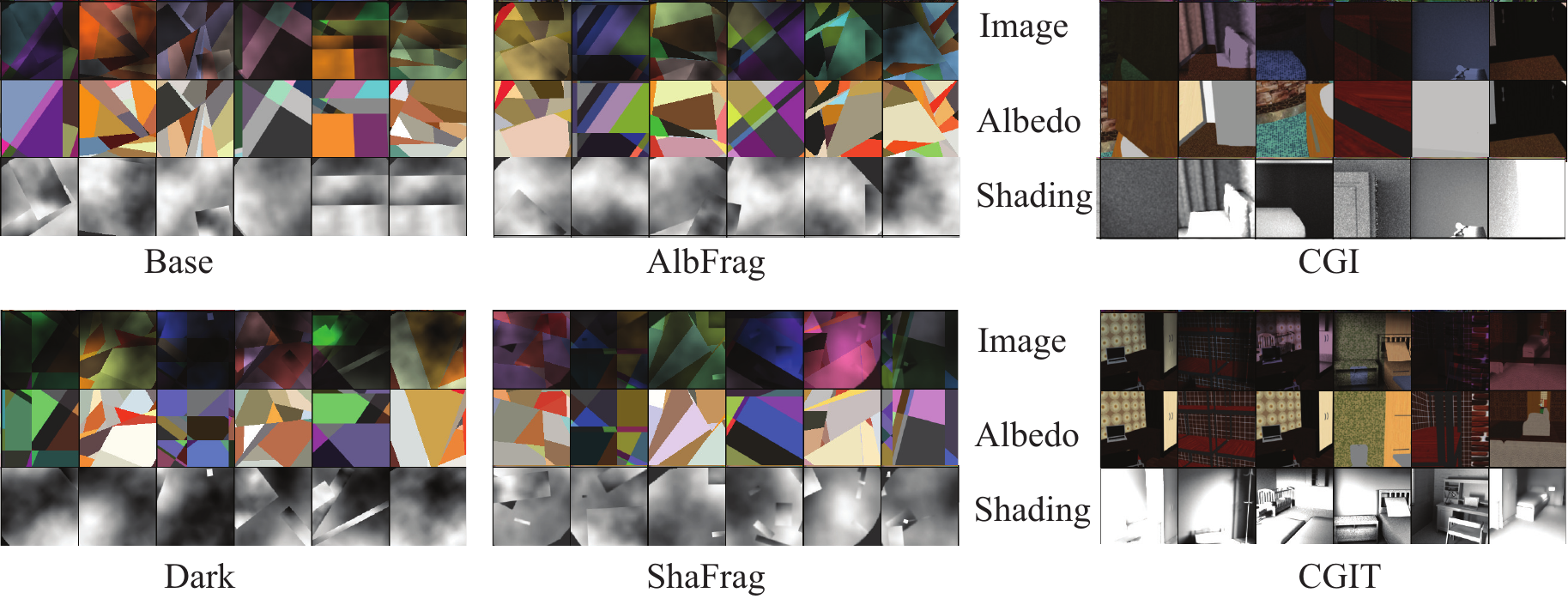}
\caption{\em Samples from our paradigms; note how there are sharp changes in
shading and slow gradients.   Generally, our albedo and shading paradigms pack more local spatial
detail into each example  compared to CGI/CGIT tiles,  but are fairly obviously not realistic.  
{\em Best viewed in color.}
}
\label{paradigmfig}
\end{figure*}

\begin{figure*}[h!]
\includegraphics[width=\textwidth]{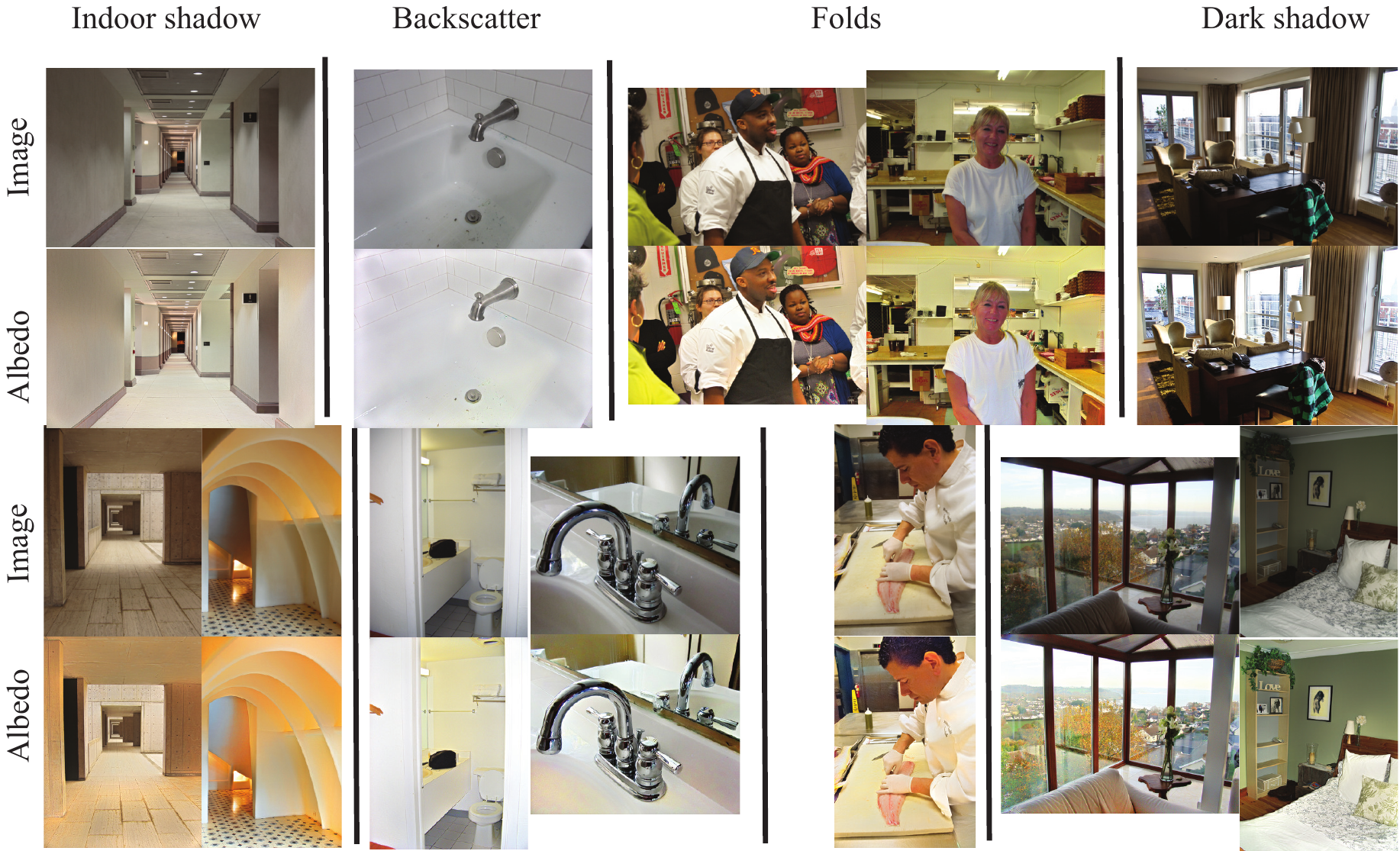}
\caption{\em Qualitative examples, from our best model ({\bf BBAF}), showing (L to R): 
  suppression of indoor shadows; suppression of backscatter from shiny bathroom
  fittings; suppression of fast shading effects from clothing folds; correctly handled dark shadow (couch back).  
\label{qualitative}
}
\end{figure*}

\begin{table*}
\centerline{
{\small
\begin{tabular}{|c|c|c|c|c|c|}
\hline
Method&Source&Training uses&Training uses& Flattening& Test WHDR\\
&&IIW labels&CG&&\\
\hline
Shi {\em et al.} '17\cite{Shi17}& \cite{EGSR18}&N&Y&N&54.44\\
Zhou {\em et al} '15\cite{Zhou:2015fp}&\cite{EGSR18}&Y&N&Y&19.95\\
*Narihira {\em et al}\cite{narihira2015regress}&ibid&N&N&N&18.1\\
Bi {\em et al} '18 \cite{EGSR18}&ibid&N&Y&Y&17.18\\
Zhou {\em et al} '15\cite{Zhou_2015_ICCV}&ibid&Y&N&Y&15.7\\
Li and Snavely '18\cite{li2018cgintrinsics}&ibid&Y&Y&Y&14.8\\
Fan {\em et al} '18\cite{8579030}&ibid&Y&N&Y&14.45\\
\hline
*Zhao {\em et al.} '12 \cite{zhao2012closed}&\cite{narihira2015regress}&N&N&N&26.4\\
*Shen and Yeo '11 \cite{Shen:2011gf}&\cite{narihira2015regress}&N&N&N&26.1\\
Yu and Smith '19\cite{yu2019inverserendernet}&ibid&N&N&N&21.4 (a)\\
 Retinex (rescaled; color/gray)&\cite{narihira2015regress}&N&N&N&19.5*/18.69*\\
*Bell {\em et al} '14 \cite{bell14intrinsic}&\cite{narihira2015regress}&N&N&Y&18.6\\
Liu {\em et al} '20\cite{Liu_2020_CVPR}&ibid&N&Y+&N&18.69\\
Bi {\em et al} '15 \cite{Bi:2015eh}&ibid&N&N&Y&18.1\\
Bi {\em et al} '15 \cite{Bi:2015eh} & \cite{EGSR18}&N&N&Y&17.69\\  
Our {\bf BBA}&&N&N&N&17.04*\\
Our {\bf BBAF}&&N&N&N&17.11*\\
\hline
\end{tabular}
}}
\caption{\em Summary comparison to recent high performing supervised (above) and unsupervised (below) methods, all
  evaluated on the standard IIW test set; sources indicated.  We distinguish
  between training with IIW and threshold selection using IIW.  WHDR values computed for Retinex use the most favorable
  scaling, using the rescaling experiments of \protect \cite{narihira2015regress}.  For our method, we report the held-out threshold value of WHDR. We report two figures for \protect
  \cite{Bi:2015eh}, because we found two distinct figures in the literature.   Key: * - method uses IIW training data to
  set scale or threshold {\em ONLY}. + - \cite{Liu_2020_CVPR} build models of albedo and shading from CGI, but does not
  use them for direct supervision.  a - \cite{yu2019inverserendernet} use patches of registered images from MegaDepth.
\label{numbers}}
\end{table*} 

\subsection{Evaluation}

Quantitative evaluation of intrinsic image methods is a recent phenomenon.  It is hard to produce data by experiment, and so only very small quantities of real albedo and
shading data are available (e.g. \cite{Grosse:2009ji,Tappen:2006vo}).  We choose to focus on WHDR measures, as they are
based on images of real scenes.
Alternative evaluations include: scores on the images of~\cite{Grosse:2009ji} (but there are very few images in
unrealistic illumination~\cite{barron2014shape}) and scores on SINTEL frames (from~\cite{Sintel}), as
in~\cite{ChenKoltun} (but this rendered data is quite unlike real images as in~\cite{Lettry}, section 2). 

The WHDR evaluation framework was put in place by \cite{bell14intrinsic}, who constructed a dataset
(Intrinsic Images in the Wild or IIW) consisting of 
human judgements which compare the absolute lightness at pairs of points in real images.  Each pair is  labelled with
one of three cases (first lighter; second lighter; indistinguishable)
and a weight, which captures the certainty of labellers.  
One evaluates by computing a weighted comparison of algorithm predictions with
human predictions; the comparison is known as the weighted human
disagreement ratio (WHDR).    Predictions were originally by testing differences in
estimated log-albedo against a standard threshold~\cite{bell14intrinsic}.  Other authors test against a threshold chosen
using validation data (eg. ~\cite{}).  Yet other authors test differences in estimated albedo (eg~\cite{}).  The choice
of predictor is significant.  Differences in log-albedo are scale invariant, but this predictor may perform poorly 
over the full range of albedos.  Two quite similar dark albedos will have the same difference in logs as two quite
different light albedos.  Differences in albedo are not scale invariant, and this means that the scale on which the
algorithm reports albedo and the test thresholds are fungible.  Some authors fix threshold, and learn scale; others fix scale and choose
threshold using validation data.   In this paper, we use differences in albedo, and test
against a variety of thresholds (section~\ref{evaluation}).  

There is a standard WHDR test-train split (20\% test and 80\% train) introduced by \cite{narihira2015regress}. The choice of scale and threshold significantly affects reported WHDR (see
table 1 of~\cite{narihira2015regress}).  Table~\ref{numbers} shows reported WHDR's for a large selection of
methods, using the best rescaled value known as appropriate.  

WHDR scores can be improved by postprocessing, because most methods produce albedo fields with very slow gradients,
rather than piecewise constant albedos.  \cite{bi20151} demonstrate the value of ``flattening'' albedo (see also
\cite{nestmeyer2017reflectance}); \cite{EGSR18} employ a fast bilateral filter~\cite{bilat} to obtain significant
improvements in WHDR.  

\begin{table*}
\begin{tabular}{|l|l|l|}
\hline
\multirow{2}{*}{Base Cases}
&{\bf Base} & all $\alpha$ are $1$; $N_t=7$, $N_\sigma=3$, average
              over 3 checkpoints.\\
\cline{2-3}
&{\bf Ma01NP}&  as {\bf Base}, but with exponential moving average during 
  training (section~\ref{adversarial}) with $w=0.9$ {\em and} \\
&&for every training pair example the decomposer sees paradigm ground
   truth\\
&& for albedo or for shading, but not both.\\
\hline
\multirow{4}{*}{Best}&{\bf BBA}& same as {\bf Ma01NP}, but $N_t=15$, $N_\sigma=5$\\
\cline{2-3}
&{\bf NP}& same as {\bf Ma01NP}, but $N_t=15$, $N_\sigma=5$ and no location code\\
\cline{2-3}
&{\bf BBAP}& same as {\bf BBA}, but with postprocessing.\\
\cline{2-3}
&{\bf BBAF}& same as {\bf BBAP}, but with discrete image averaging as
             well.\\
\hline
\hline
\multirow{5}{*}{Variant Smoothing/Averaging}
&{\bf NoSmo}& (no adversarial smoothing), $\alpha_d=0$.\\
\cline{2-3}
&{\bf NoInt} & (no interpolation), $\alpha_a=\alpha_s=0$ (so all
               training comes from the adversarial smoothing).\\
\cline{2-3}
&{\bf NoRes}& (no residual), $\alpha_{rr}=0$, no residual loss for real examples; this fails utterly
             (Figure~\ref{summary}).\\
\cline{2-3}
&{\bf BBAT}& same as {\bf BBAP}, but with discrete tile averaging as
             well.\\
\hline 
\hline
\multirow{4}{*}{Variant Discriminator Scales}
&{\bf SD}& $R_d=10$, otherwise as {\bf BBAF}\\
&{\bf ID}& $R_d=29$, otherwise as {\bf BBAF}\\
&{\bf MD}& $R_d=48$, otherwise as {\bf BBAF}\\
&{\bf BD}& $R_d=128$, otherwise as {\bf BBAF}\\
\hline
\multirow{6}{*}{Variant Paradigms}
&{\bf CGI}&  albedo and shading tiles from CGIntrinsics ~\cite{Li:2018if} are
           used rather than paradigm images.\\
&& Tiles are selected from shading and albedo independently.\\
\cline{2-3} 
&{\bf CGIT}& albedo and shading tiles from resized versions of
             CGIntrinsics ~\cite{Li:2018if} images are used\\
&& rather than  paradigm images; resizing is to 180 pixels on the shortest edge, and ensures albedo
  tiles \\
&&have more structure; tiles are selected from shading and albedo
   independently.\\
\cline{2-3} 
&{\bf CGITD}& albedo and shading tiles from resized versions of
              CGIntrinsics ~\cite{Li:2018if} images are used\\
&& rather than
  paradigm images;
resizing is to 180 pixels on the shortest edge, and ensures albedo tiles\\
&& have more structure; dependence between shading and
  albedo tiles is preserved.\\
\cline{2-3} 
&{\bf Dark}&  the paradigm for shading is modified to have a higher
             dynamic range ($s_{\mbox{min}}=0.05$).\\
\cline{2-3} 
&{\bf AlbFrag}& the albedo paradigm contains very small fragments;
                $d_{\mbox{max}}=9$, $p_{\mbox{min}}=100$.\\
\cline{2-3} 
&{\bf ShaFrag}& the shading paradigm contains very small fragments; $n_m=16$.\\
\hline 
\hline 
\end{tabular}
\caption{\em Key to models trained and evaluated.  Note that {\bf Ma01NP}, {\bf BBA}, {\bf BBAP}, {\bf BBAF}, {\bf BBAT}
  differ only by inference procedure (all use the same generator network parameters).  Other models are trained using
  different data, losses or discriminators..
\label{key}
}
\end{table*}

\subsection{Supervision}

Direct supervision occurs when a method sees the albedo and shading of training images.  
With even a few ground truth images are available, local regression strategies have been successful 
\cite{Tappen:2006vo}.  The recent literature strongly emphasizes
directly supervised CNN based models. One option is to \cite{narihira2015direct} regress lightness differences
against image features using IIW data.  \cite{Zhou_2015_ICCV} smooth pairwise lightness comparisons (learned using WHDR
data) to albedo and shading fields using a fully connected CRF.  Recent methods emphasize direct supervision using CGI
rendering of scene models~\cite{ChenKoltun,narihira2015direct,Li:2018if}
However, models trained exclusively on rendered scenes do not do well on real images (eg~\cite{Lettry}; section 2).
This is likely because rendered images are insufficiently ``like'' real images in some important ways.  Competitive modern methods are trained using a
training portion of the IIW dataset, then evaluated on a the test portion. \cite{8579030} obtain a SOTA WHDR of
14.45\% in this way, but their method produces strange colors in albedo images, making its applicability in
computational photography questionable and qualitative comparison unhelpful. \cite{EGSR18} use a similar approach, but different network architectures, to
obtain a mean WHDR of 17.18\% with strong qualitative results; we use this method for qualitative comparison.
There is good evidence that relatively little supervision is required, and that self-supervision can be successful.
\cite{Janner} apply a learned renderer to decompositions of unlabelled data to obtain a residual loss that improves 
performance.  \cite{Cheng_2018_CVPR} show that a form of bootstrapping (augment training data with the results of previous models)
is effective in improving performance.  

Indirect supervision occurs when a method does not see the albedo and shading of training images, but
sees equivalent information.  This can take a variety of forms. 
Aligned views of the same real scene under distinct illuminants offer strong cues
to intrinsic image decomposition, exploited  in~\cite{Weissseq,BigTimeLi18,Laffont_2015_ICCV}.  Alternatively, one
can use aligned CGI renderings of the same scene~\cite{Lettry}.  
 \cite{ma2018single} show
how to exploit these cues to learn a method that, at inference time, can be applied to a single view. 
\cite{yu2019inverserendernet} show that it is enough to partially align images of real scenes (by 
matching sections of frames).   

Indirect supervision can take a more abstract form by providing the method only statistical models of albedo
(resp. shading), much like the original Retinex assumption.  \cite{Liu_2020_CVPR} 
who use albedo and shading CGI renderings to build autoencoders.  These are used to 
impose albedo (resp. shading) structure on the inferred components of the input image; the components must
also compose to make the image.  This method obtains the current SOTA WHDR for methods that use only
indirect supervision (18.69\%).    Our method also receives only statistical models of albedo and shading, but it 
receives them directly.  We multiply samples from albedo and shading paradigms, and train the method to decompose
the product into the original samples.  This training data is quite unlike real images or CGI renderings, and
we rely on adversarial smoothing to ensure the decomposer applies to real images, resulting in a new SOTA WHDR  for
indirect supervision (17.04\%).   

\ignore{\subsection{Image Regression and Adversarial Smoothing}
\todo{CGAN comparison? also, cycle gan, im-2-im, etc, patchgan}}

\subsection{Invariance and Equivariance}

Most applications must control how a CNN behaves when an image is transformed.  A classifier, for example, should not
change prediction if the image is shifted or scaled.  There is no crisp theoretical 
framework for transformations of the input.  The theory of group actions does not apply exactly to image rotations,
scaling or cropping,  because almost all interesting transformations of this form involve information being
gained or lost at the boundary of the image.    For image classification, data augmentation --- training with
multiple crops, scalings, colorings and rotations of training examples --- seems to result in classifiers that are
robust to transformations (origins uncertain; survey in~\cite{augsurvey}).  Averaging predictions over multiple distinct
crops is now universal practice (origins again uncertain).  Augmentation and averaging result in a property analogous to
invariance, though a precise definition remains obscure (early attempts, in another context, in~\cite{qi1,qi2}).
Imposing augmentation robustness seems to constrain a representation strongly. 

More important in regression applications is equivariance.  A function $\phi: {\bf x} \in X \rightarrow {\bf y} \in Y$ is equivariant under the action of a group $G$ if there
are actions of $G$ on $X$ and $Y$ such that $\phi(g \circ {\bf x})=g \circ \phi({\bf x})$.  Again, information being
gained or lost at the boundary is an obstacle to applying the theory of group actions exactly (except for certain finite
groups~\cite{CohenWelling}).   If one relaxes the definition to require only an approximate match, well-known visual
feature representations tend to have strong equivariance properties either by design or in practice~\cite{LencVedaldi}.
Generally, equivariance properties have not been imposed on regression networks; we know of no better strategy
for doing so than averaging. 

\todo{Anand's papers on padding}

\section{Framework}

We model an image as a colored albedo field multiplied by a shading field and a single color.  Generally, we
use bold for vectors (position, color fields) and so write $\im(\vect{x})=\al(\vect{x}) \circ \sh(\vect{x})=\al(\vect{x}) \circ \left[ \shm(\vect{x})\vect{c}\right]$
where $\vect{c}$ is the color of the shading field and $\circ$ is elementwise multiplication.   

There is strong support in the literature for the model of albedo as patches of constant color.  For example,
postprocessing with the fast bilateral filter makes this assumption and is helpful~\cite{EGSR18}; most priors 
are derived from this assumption; most current regression methods produce albedos that look like patches of constant
color (eg Figure~\ref{multiqual}).    We adopt this model.  Our shading model supports more complex phenomena like fast shading
edges (like the cast shadows or the cloth folds in Figure~\ref{qualitative}).  Imposing these models poses what are
essentially local problems such as deciding how an image gradient should be decomposed into shading and albedo effects;
once these are solved, the albedo can be determined by ``filling in'' appropriate constant colors.   But it is
inconvenient to determine the details, or to (say) optimize a posterior.  Instead, we  train a fully convolutional network directly on
synthetic examples which can represent how the local problems are to be solved (section~\ref{paradigms}); we then use adversarial smoothing
methods to ensure that the network produces reasonable results on real images (section~\ref{adversarial}).  Our network could be applied to any size
of image, because it is fully convolutional; but doing so ignores the significance of scale in intrinsic image
problems and produces solutions without the required equivariance properties.   Instead, the network is trained on fixed
size tiles, and the results on tiles are reassembled into a (somewhat) equivariant estimate (section~\ref{covariance}).  Finally, the result is
postprocessed per-pixel to ensure that albedo and shading compose to make the image (section~\ref{postprocessing}).

\subsection{Paradigms}
\label{paradigms}

Our synthetic albedo (resp. shading, color) models, paradigms in what follows, 
are samples from easily sampled random processes that produce tiles that appear to capture the important properties of
albedo (resp. shading, color) at a short scale.  Paradigms can be thought of as priors represented
in a form that is convenient -- rather than a loss that depends on the prior, we train the
network to decompose examples from a prior model.  For some kinds of constraint -- for example, the requirement
that an albedo be piecewise constant, with sharp edges -- there may be a practical advantage to representing the prior
with paradigms, rather than as a cost function, because it can be difficult to author cost functions to capture these
constraints accurately.  We require that paradigms represent albedo and shading only on a relatively short scale.   This
means that paradigm samples do not need to look like real albedo (resp. shading) images. The paradigms must be chosen by
hand (we have no search procedure for paradigms).  

Our albedo paradigm uses a surface color model and a spatial model.  The qualitative properties it is intended to capture are:  albedoes are piecewise constant; the color distribution
should reflect likely surface colors; there should be a profusion of edges with no strong orientation bias; there should
be at least some vertices with degree greater than three.   Surface color is modelled by drawing
color samples uniformly and at random from the IIW training set.  These must be adjusted for presumed illumination.  We
do so by assuming the range of illumination intensity is approximately the same as the range of lightnesses,  and so
dividing by the square root of intensity.  

The spatial model is an evenly weighted mixture of two spatial models. The first models the albedo as a kd tree, with
spatial splits chosen at random, a fixed maximum depth ($d_{\mbox{max}}=6$ unless otherwise stated), and a fixed minimum number of pixels per
cell ($p_{\mbox{min}}=1000$ unless otherwise stated).  For each cell, the color is chosen uniformly at random from the surface color model.   
The second models the albedo as a mondrian of rotated mondrians.  We build a dictionary of rotated mondrians by first 
constructing random axis aligned rectangular grids, then filling in each grid cell with a sample from the
color model, then applying a random rotation.  Each mondrian is then obtained by constructing a random axis aligned
grid (of $n_c$ cells on edge), and filling each grid cell with a correspondingly sized, randomly selected block from a
random dictionary entry.   The number of cells on edge $n_c$ is chosen uniformly and at random in the range $1$ to
$n_m=4$ (unless otherwise noted).

Our shading paradigm uses a spatial model to combine samples from Perlin noise.  The qualitative properties it is intended to
capture are:  shading contains many slow and very slow changes; there are some sharp shading edges; and  the dynamic range of
shading indoors is limited.    Our shading model uses a Perlin noise field, constructed from five scales of smoothing.  We
construct five dictionaries, one per scale ($\sigma \in \left[3, 6, 12, 16, 24\right]$).  Each dictionary contains IID
unit normal images smoothed with gaussians at the corresponding $\sigma$.   A shading component consists of a weighted
sum of randomly chosen elements, one per dictionary, weighted by $\left[0.2, 0.2, 0.4, 1, 1\right]$ respectively.   
A shading sample is obtained by: randomly constructing a shading component as a background; choosing a random number of masks to impose; then, for
each mask, replacing the shading in the interior of the mask with the shading from another, randomly chosen, shading
component.   The masks are chosen from two options: leaves in a random kd tree of fixed maximum depth ($d_{s\mbox{max}}=6$) and minimum number of
pixels per leaf ($p_{s\mbox{min}}=1000$); or cells in a dictionary of rotated mondrians, constructed as per the 
albedo mondrians.  The resulting sample is rescaled to have fixed minimum ($s_{\mbox{min}}=0.2$) and maximum ($s_{\mbox{max}}=1$)  value.  
Figure~\ref{paradigmfig} shows typical samples.  

Our model assumes there is no spatial variation in illumination color.  A sample from the color paradigm is given by
$0.5*\left(1, 1, 1\right)^T+0.5*\xi$, where $\xi \sim N({\bf 0}, {\cal I})$.  This means that paradigm images can have
quite strong color casts (Figure~\ref{paradigmfig}).

For synthetic examples, we know: image; albedo; color; and shading.  We must ensure that the predicted albedo for
the tile is close to the true albedo (${\cal L}_a$); 
the predicted shading is close to the true shading (${\cal L}_s$); the
predicted color is close to the true color (${\cal L}_c$); and the 
image is explained by the albedo and shading (${\cal L}_r$).  We work with
images, rather than log images, so albedo and shading must multiply to
yield the image.  Our loss is 
\[
{\cal L}_T(\theta)=\alpha_{a} {\cal L}_{a}+\alpha_{s}{\cal
  L}_{s}+\alpha_c {\cal L}_c +\alpha_r {\cal L}_{r}
\] (details in appendix~\ref{losses}).

\begin{figure*}
\includegraphics[width=\textwidth]{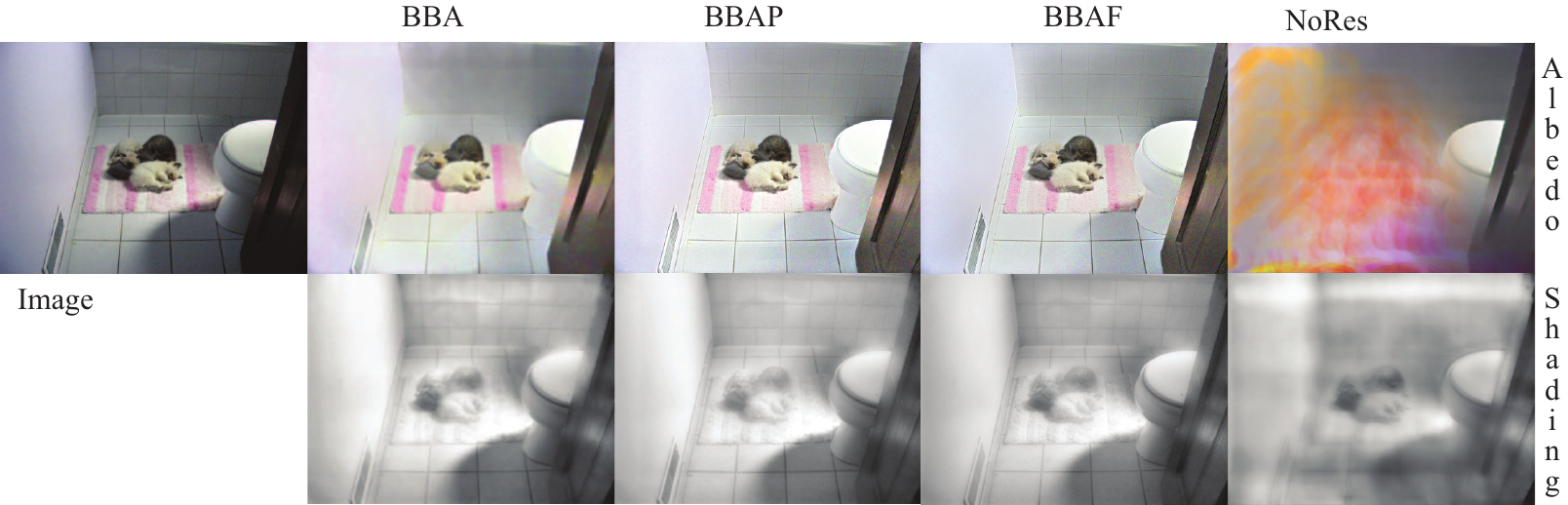}
\caption{\em  Our strongest models are very strong, and differ in
  minor details. The slightly blurry albedo of {\bf BBA} is made sharper
  by postprocessing ({\bf BBAP});  postprocessing, by forcing albedo
  and shading to have very small or zero residual, reduces the slight
  color error of the left wall.  Qualitative differences between {\bf BBAP} and
  {\bf BBAT} are (typically) very hard to spot, though note the
  slightly sharper shadow boundaries and the slightly deeper tile
  grooves.  Finally, {\bf NoRes} fails so catastrophically there was
  no point in continuing training (which is why no numerical results
  are reported for this model).
{\em Best viewed in color.}
\label{summary}
}
\end{figure*}

\begin{figure*}
\includegraphics[width=\textwidth]{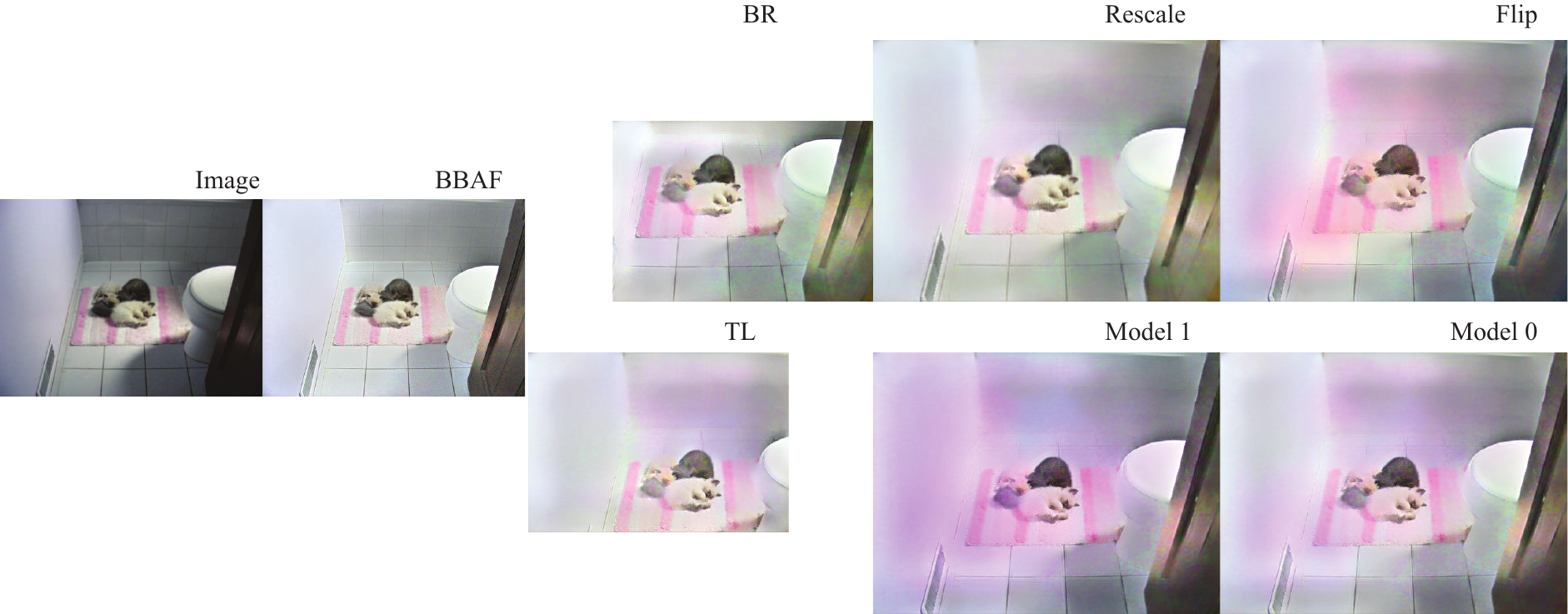}
  \caption{\em A naive application of our adversarial smoothing procedure works poorly -- equivariance failures are severe
    and punishing.  For the image shown we compare albedo reconstructions from a reference model ({\bf BBAF}, our best)
    with others.  In this case, each image has been passed through the underlying model (which is convolutional, and so
    applies to any scale).  {\bf Model 1} and {\bf Model 0} are different checkpoints, separated by approximately 10,000
    training images; notice how there are significant long scale differences, caused by the fact that the adversarial
    smoothing does not identify a unique best model.  As Figure \protect \ref{smoothing} shows, an exponential moving average resolves this
    effect.  {\bf Scale} shows the (rescaled) albedo for an image that was rescaled down by 1.4, then decomposed using
    model 0.  Comparing this to the result of model 0 shows a severe failure of scale equivariance.  Similarly, {\bf Flip}
    shows the (reflected) albedo for an image that was reflected in both axes, then passed through model 0; comparing
    this to the 
result of model 0 shows a severe failure of rotation equivariance. Finally, {\bf BR} and {\bf TL} show the results of
cutting the image into two overlapping tiles, and passing each through the network; comparing these shows a severe
failure of translation equivariance.  The symptom of these equivariance failures is long spatial scale error of a form 
disruptive to WHDR comparisons.  Our strong WHDR performance shows that our averaging procedures control these effects.
    \label{oscillationfigure}
  }
  \end{figure*}

\subsection{Adversarial Smoothing}
\label{adversarial}

For real examples, we do not know albedo or shading, but we can ensure that the image is explained by the
albedo and shading (${\cal L}_{rr}$), and predicted albedo and shading are within a reasonable range (${\cal L}_{\mbox{range}}$).  
Our loss is
\[
{\cal L}_R (\theta)=\alpha_{rr} {\cal L}_{rr}+{\cal L}_{\mbox{range}}
\]
(details in appendix~\ref{losses}).
But this loss does not control what the model does to real tiles in any detail.  Here is one way to diagnose whether
the model is mapping real tiles appropriately.  Take a population of real tiles, and decompose
them.  Cut pairs of patches of some appropriate fixed size out of each of the resulting albedo and shading 
fields -- call these the real data pairs.  Similarly, cut pairs of patches of the same 
size from the training data -- call these the training data pairs.   Because albedo (resp. shading) has a local character,
we expect that, if the patch size is sufficiently small, real data pairs should be ``like'' test data pairs; any reliable
distinction between the two categories is a sign that the model may not be behaving properly.    We do not
seek to match the distribution of albedos for decomposed images to that of paradigms (which doesn't work particularly
well, as shown in Figure~\ref{scalefx}).  Instead, we impose distribution matching only at the scale of patches, 
because we can trust the paradigm model only at fairly short scales.

For the real pairs, the decomposer is a generator, because it makes image pairs for which we
know no loss, so we can use an adversary to refine it.  Some modifications are required. 
It is usual to write an adversarial loss and seek a saddle point~\cite{GoodfellowARXIV2014}.  
If the saddle point exists (unlikely; see~\cite{pmlr-v70-arora17a}),  the generated 
distribution matches the data distribution \cite{GoodfellowARXIV2014}.  For our purposes, this matching may be
undesirable, as the training pairs may be at best a rough approximation of what real pairs look like.   In practice,
generators are implemented by taking some steps on the discriminator for fixed generator, then some steps of the
generator for fixed discriminator.  We follow this procedure (details in Appendix).

In this case, training dynamics may not converge to a single model,
but rather wander around a stationary set of distinct models each of which produce
somewhat different reconstructions.  This effect may not be a nuisance for generators because  
it affects only the mapping from latent variable to image, which doesn't usually matter. In our case, it is a potentially serious nuisance, because
different checkpoints taken at the end of training may report very different albedos for the same image (Figure~\ref{oscillationfigure}).
We manage this effect by averaging model parameters, either over a fixed number of checkpoints or using a moving
average of parameters,  which gives better results.  Write $\theta$ for the current estimate of the generator
parameters; we maintain a separate set of parameters $\psi$, and update them by $\psi\rightarrow w \psi+ (1-w) \theta$
every 5000 images.

\begin{figure*}
\includegraphics[width=\textwidth]{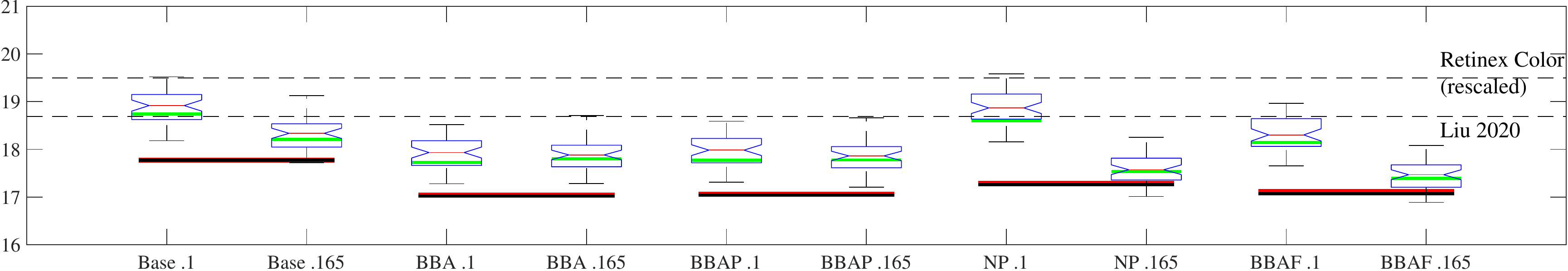}
\caption{\em Our best methods are strong. {\bf Base} outperforms Retinex
  Color and \protect \cite{Liu_2020_CVPR}, and note near .17 WHDR with
  held out threshold for {\bf BBA}, {\bf BBAP}, {\bf BBAF}).  The standard WHDR
  test set may be easier than most subsets that size (green bars well below median in boxplots).  Postprocessing and
  flipping may  appear to weaken performance (cf red/black bars for {\bf BBA}, {\bf BBAP} and {\bf BBAF}), but this is an artifact of using one
test set; as Figure \protect \ref{treat} shows, {\bf BBAF} beats {\bf BBA} and {\bf BBAP} for almost every simulated test set.  
{\bf Key:} {\em Fixed thresholds:} shown in boxplots of WHDR values for 50 simulated test sets for the two fixed
thresholds, and green bars are the value for the standard test set.  {\em Oracle thresholds:} heavy black bar.  {\em
  Held out threshold:} heavy red bar. {\em Boxplots:} horizontal bar = median; notch = fraction of interquartile
range outside which a difference in medians is significant; bottom and top of the box = 25 and 75 percentiles resp.;
whiskers extend to the most extreme data points that are not outliers; outliers -- greater than 1.5 times the
interquartile range outside top and bottom -- are '+'. 
{\em Best viewed in color.}
\label{best}
}
\end{figure*}

\subsection{Equivariance and Averaging}
\label{covariance}

An ideal intrinsic image method will report the same albedo for the same location in a scene, however that location is
viewed.  We know no way to impose this criterion.  A simpler equivariance requirement is that all image tiles (however located, oriented or scaled)
containing some point $x$ in the scene will report the same albedo and shading for that point.  
Note first that there is a problem to solve here: even a fully convolutional network is not
equivariant under shifts of the image, because of  boundary effects -- some locations in the output depend in some way
on units whose  support extends  outside the image and into the padding.    This means that a pixel in the overlap of
two tiles could be estimated (say) with padding in the first estimate and without padding in the second, and different
estimates will result.  A natural way to impose this equivariance requirement is to estimate the albedo at each point as the average
of estimates made by multiple tiles (with different offset, location
and scale) containing that point. 

Experimental images are approximately 400 pixels on edge, with some  range of variation.   Cropping tiles of
arbitrary scale and orientation is inefficient.  Instead, for each scale, we average over a random set of tiles of fixed size.
At a given scale, we cut images into a $N_t \times N_t$ grid of overlapping tiles, with dithered centers, arranged to
cover the image, and then form a weighted average of the results for the tiles.  Tiles are organized to ensure that each pixel is
covered by at least one tile, though most pixels are covered by many tiles. We use a weighted average to suppress ringing artifacts; weights decline exponentially to the
boundary of the window (detailed form in appendix).  We have not experimented with other window forms.   To ensure that feature computation takes into account whether a
location is near the center of the tile or near the edge, we augment input tiles with a simple location code (detailed
form in appendix). We have not experimented with other location codes. 

We average the albedo and shading estimates so obtained for several rescaled versions
of the original image, and average.  We average translation averaged albedo and shading reconstructions over $N_s$
scales spaced evenly from approximately $1/\sqrt{2}\times \mbox{image size}$ to $\sqrt{2}\times \mbox{image 
  size}$.  

It is trickier to achieve equivariance under orthogonal transformations
by averaging. Recall that an orthogonal transformation is a rotation
possibly composed with a reflection.   The number of samples required becomes large, and
extracting tiles (resp. images) at arbitrary rotations is inefficient.
We have investigated two simplified strategies.  In the first, we
compute all eight images obtained by rotation by a multiple of $90^0$
composed with a reflection, compute translation and scale averaged decompositions for each, then
average the results ({\em discrete image averaging}).  In the second, we average over
all eight tiles so obtained for each tile processed {\em during} scale
and translation averaging ({\em discrete tile averaging}).    These averaging steps increase inference time eightfold, and so
we investigate their effects only for models known to be strong.

\subsection{Averaging Controls Error}

Averaging across scale, translation and rotation helps control some form of model error.
 The smoothing procedure ensures that a generic image will produce an albedo (resp. shading) field that ``looks like''
the training data {\em at the scale of patches}.   The albedo is not controlled on a longer scale.  This means that
the predicted albedo for a tile may contain an error that is on a longer scale than the size of a patch and that depends on the input image.
Here are some examples.  The albedo model consists of piecewise constant patches, and the shading model contains some
fast shading boundaries (section~\ref{paradigms}).  The network could predict a fast change in albedo coordinated with a
fast change in shading.  Alternatively, the network could predict an albedo that has a slow (but not zero) gradient that
is low enough that the difference from zero is hard to resolve at the scale of a patch. 
This error may depend on the long-scale structure of the input image -- for example, mostly red images might get
spurious fast changes in albedo.   One strategy to control this effect is to have models that are very good on long
spatial scales, too; but we do not know how to produce training data that properly represents the desired outcome at a long scale.

The error in each tile will be referred to the coordinate system of that tile.  As a result, for any given location $x$,
we are averaging estimates of albedo and shading that have different error terms (because they have different locations
in the tile coordinate systems for their tiles).  Equivalently, the context used to produce the estimate at $x$ is
different from tile to tile.    As a result, we expect the averaging process to suppress errors at long spatial scales.
Figures~\ref{oscillationfigure} and~\ref{smoothing} strongly suggest that this error control is important. Discrete image
averaging significantly outperforms discrete tile averaging, likely because discrete tile averaging cannot control error
on long scales.

An alternate view is this.  Each albedo (resp. shading) in a tile estimate is the result of an estimator (the function
implemented by the network at that point).  But not every estimator is the same; some have support that reaches into the
padding.  Training ensures that the expected error of each estimator is zero, or close, but does not ensure that
estimators have the same variance.  By shifting, rotating, and scaling images, we are essentially producing multiple
distinct estimates of the same albedo (resp. shading), and averaging reduces their variance.

\subsection{Postprocessing}
\label{postprocessing}

Averaging at a fixed scale has two important effects.  First, the color estimate $\vect{c}$ is no longer constant as a
function of position (each tile produces a constant, but the average may not be).  Second, averaging means that the
residual might be larger than desired.  In particular, fine details in the images may be obscured by averaging across
scales.  These effects can be fixed at inference time by post processing, and the results demonstrate a small advantage
to doing so (Figure~\ref{best}).  Assume that $\im$ has produced averaged albedo estimate $\al(\vect{x})$, averaged shading
estimate $\shm(\vect{x})$, averaged color estimate $\vect{c}(\vect{x})$ and residual $\vect{r}(\vect{x})=\im-\al\circ
[\shm \vect{c}]$.  Then we seek small $\delta \al(\vect{x})$, $\delta \shm(\vect{x})$ so that $(\al +\delta \al) [(\shm +\delta \shm) \vect{c}]$ is closer to
$\im$.  As the appendix establishes,
  \[
\delta s=\frac{\vect{r}^T\vect{a}}{\vect{a}^T\vect{a}+s^2} \mbox{    and   } \delta \vect{a} =
(1/s) (\vect{r}-\vect{a}\frac{\vect{r}^T\vect{a}}{\vect{a}^T\vect{a}+s^2})
\]
 Note that (a) the process can be iterated and (b) the computation is pointwise and fast.  Our experience has been that
 the averaging method produces a fairly small residual, and few iterations are required. 
Where noted, postprocessing is applied for each scale's average, and then for the average across scales.

\subsection{Network Details}
\label{discscale}

{\bf Discriminator network:}   We want our discriminator score to depend only on local neighborhoods.  We use a straightforward trick, derived from the
practice of training adversarial networks using a hinge loss~\cite{}. Write $\im$ for the input field, 
$y$ for the label (-1 for real, 1 for generated). We achieve a local discriminator by structuring the network
as a set of convolutional layers to produce a $1\times k \times k$ tensor ${\cal F}$ from $\im$.   The scale of the
patches is dictated by the size of the receptive field for the elements of ${\cal F}$.  The discriminator is trained by
using a mean hinge loss over all overlapping patches of that scale; this can be computed by computing
$\mbox{mean}(\mbox{ReLU}(1-y {\cal F})$.     The loss used to train the generator is then obtained as $\mbox{mean}({\cal F})$.
 All experiments, except scale experiments,  use the same network structure
for the adversarial discriminator (Appendix).   Scale experiments vary the number of layers and the size of the
kernel to achieve the given patch size.    All discriminator networks are trained with leaky ReLU's and spectral normalization.

{\bf Decomposer network:} Our network accepts a $128\times 128$ image tile $\ipt$ and produces a $3 \times 128 \times 128$ estimate of absolute
spectral albedo $\iat(\ipt; \theta)$ (i.e. our albedo estimate is colored), a $1 \times 128 \times 128$ estimate of
absolute shading $\ist(\ipt; \theta)$, and a $3$ dimensional estimate of
illuminant color $\vect{c}(\ipt; \theta)$.  All experiments use the
same network structure (Appendix), but with different losses and different training data as noted.  

{\bf Training details:} All experiments
train the network in single precision and use a batch size of 128; all networks see a total of 32M training images, evenly
divided between real tiles and paradigms.  For all experiments, training albedos for
a batch are selected uniformly and at random from a cached dictionary of 4000 samples; similarly, training shading is
selected uniformly and at random from a cached dictionary of 4000 samples.  Real tiles are selected uniformly at random
from a dictionary of 4000 samples.  These samples are drawn only from the standard training set for IIW.

\begin{figure*}
\includegraphics[width=\textwidth]{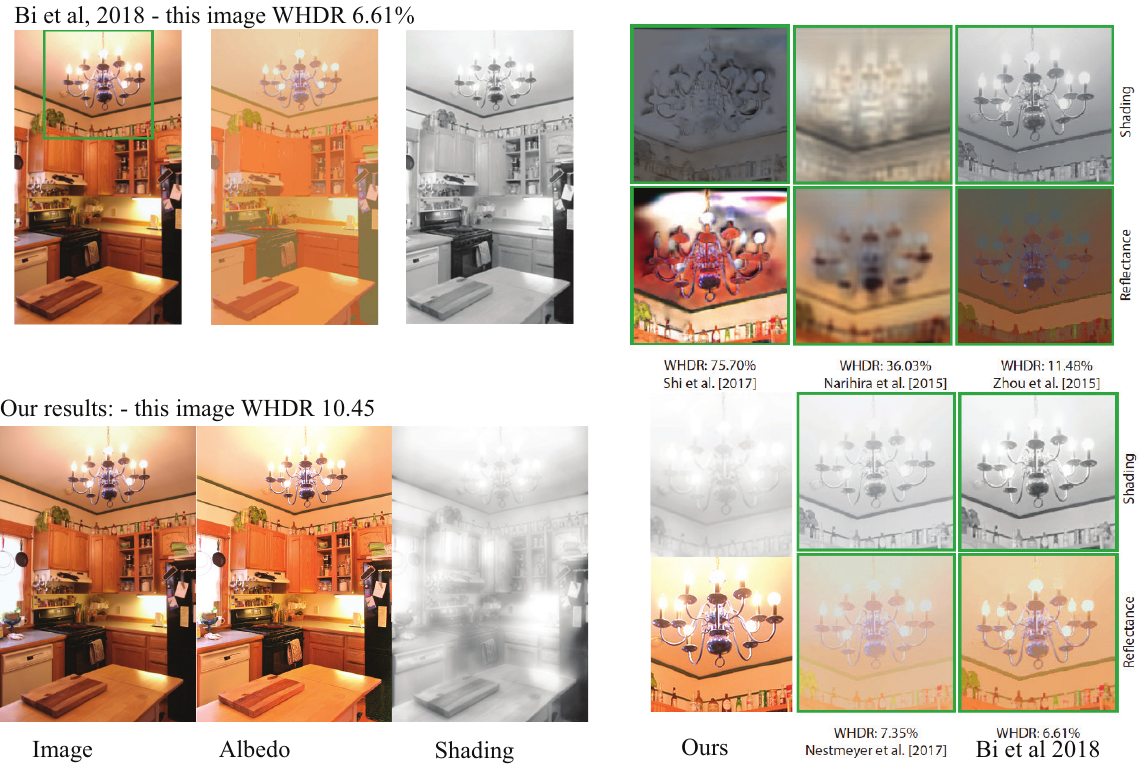}
\caption{\em Qualitative comparison to \protect \cite{EGSR18},
  \cite{Shi17}, \cite{narihira2015direct},
  \cite{nestmeyer2017reflectance} and \cite{zhou2015learning}, using
  parts of Figure 1 of~\cite{EGSR18}.  As \cite{EGSR18} remark,
  the methods of \cite{Shi17} and \cite{narihira2015direct} are
  trained on rendered data alone, and face difficulties due to the
  difference between rendered data and real images.  As \cite{EGSR18}
  remark,  the methods of \cite{narihira2015direct} and 
  \cite{nestmeyer2017reflectance} face difficulties due to the deep 
shadows in the scene.  The albedo produced by our method does not show
the ``colored paper'' effect seen in other methods and does not
produce odd colors; this is an advantage (text).  
Our method reports
  albedo and shading up to image boundaries, that of \cite{EGSR18}
  appears not to (the crop of the figures is as in the original paper;
  for our method, we show the whole image).  
\label{multiqual}
}
\end{figure*}

\section{Evaluation}\label{evaluation}

Our procedure is intended to produce colored albedo (surface color) estimates.  
We reduce these to lightness estimates by averaging the three color channels, and test the difference in predicted
lightnesses against a threshold.  If the absolute  chosen in one of three ways.  For comparison with other algorithms, we compute WHDR on
the standard test set, using both a {\bf held-out threshold} (chosen as the threshold that gives the best WHDR for
the training set) and an {\bf oracle threshold} (the threshold that yields the best WHDR). 
Because  we wish to investigate the performance of lightness algorithms that have never seen real training data, we evaluate {\bf fixed
  thresholds} (chosen in advance and largely independent of the WHDR dataset) are our primary interest.  
We investigate two thresholds: 0.1 (because Bell \ea \cite{bell14intrinsic} used this value, and because this yields about 10 
distinguishable lightness values) and 0.165 (because a search on WHDR validation data gives this threshold as the one at
which the differences in image intensities yields the best WHDR on a validation set; this is the only reliance on IIW
data in choosing this threshold). 

{\bf Models:} We have investigated a number of models.  
Models use variants of our loss, summarized below for convenience. Paradigms are used as training data; all $\alpha$ are
$1$; $N_t=7$ and $N_\sigma=3$ except where explicitly noted. Table~\ref{key} lists models.

\begin{figure*}
\includegraphics[width=\textwidth]{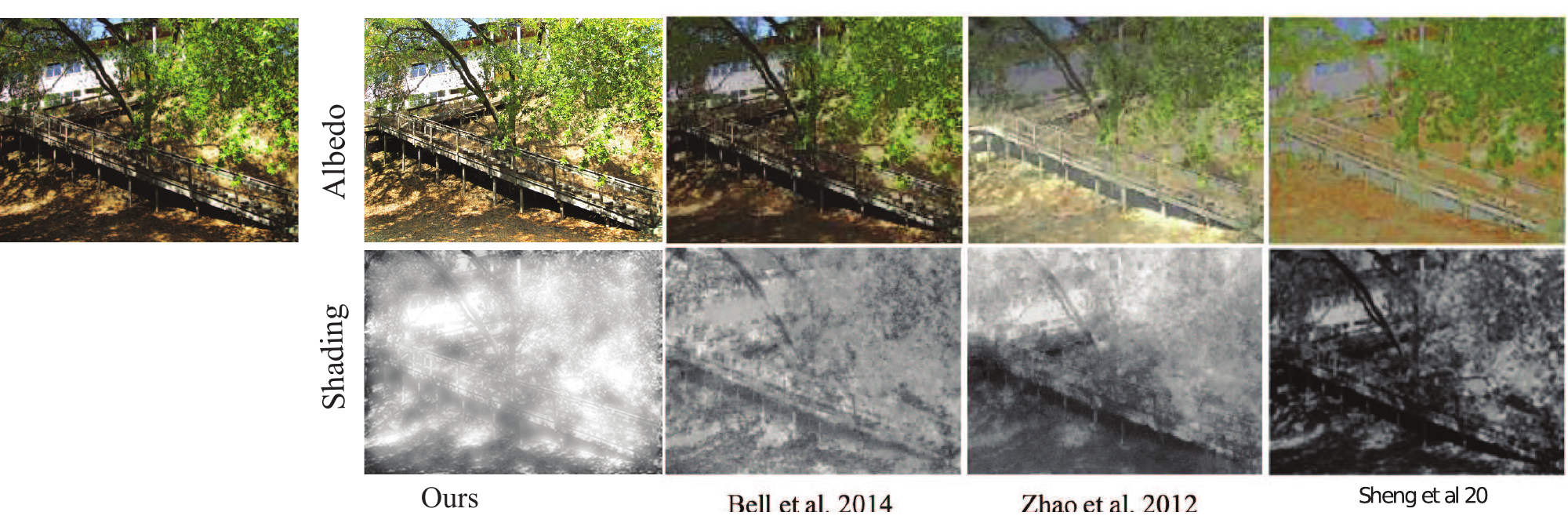}
\caption{\em Qualitative comparison to \protect
  \cite{bell14intrinsic},~\cite{zhao2012closed} and~\cite{Sheng20},
  using in part Figure 14 of~\cite{Sheng20}.  The method of \cite{Sheng20} is
  more successful than others at suppressing this complex mixed shadow, but
  produces ``colored paper'' effects in the albedo. The method
  of \cite{bell14intrinsic} does not handle the shadows well; the
  method of \cite{zhao2012closed} is better, but washes out the
  albedo.  By comparison, our method is moderately successful on this
  challenging image. 
{\em Best viewed in color.}
\label{shengcompare}
}
\end{figure*}


\subsection{Standard Test WHDR}               
\label{whdrcompare}

Other published methods are allowed to see training WHDR data (Retinex does so to choose a scale), so 
the our methods can be compared using WHDR on the standard test using the held out threshold.  Here our method sees the
training set to choose threshold (but for nothing else).   As table~\ref{numbers} indicates,
in this comparison, our best method ({\bf BBA}) with 17.04\% WHDR strongly outperforms other unsupervised methods and
is comparable to recent strong supervised methods. However, this comparison is not a particularly good way of choosing models.

\begin{figure*}
\includegraphics[width=\textwidth]{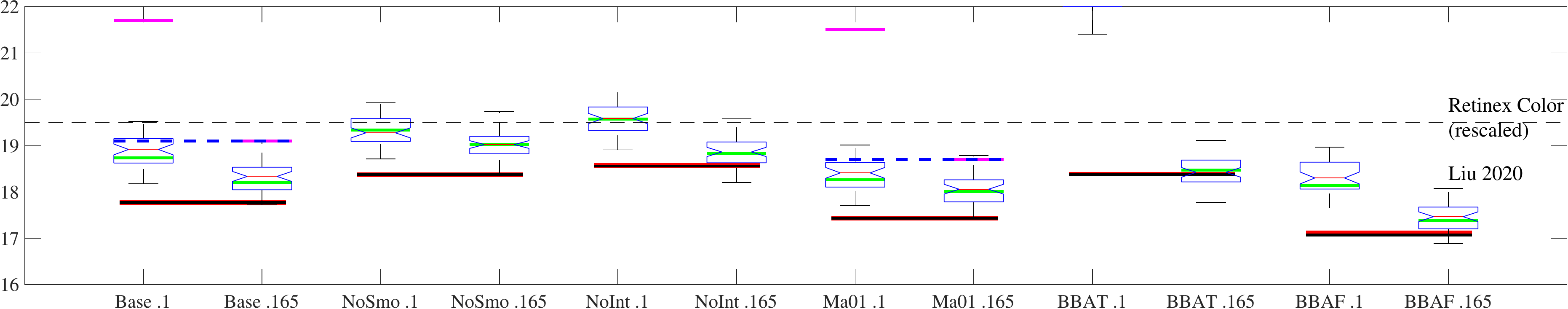}
 \caption{\em Smoothing, averaging and postprocessing are important.  Without adversarial
  smoothing ({\bf NoSmo}), performance is comparable to Retinex.  Adversarial smoothing {\em alone} ({\bf NoInt}) is
  surprisingly well behaved.  Averaging makes a very significant difference (compare blue/black bars and purple/green
  bars) and averaging over a larger number of tiles is better (cf.
  {\bf BBA} and {\bf Base}).  Discrete image averaging results in  improvements (cf. {\bf BBA} and {\bf BBAF}), and is
  clearly better than discrete tile averaging (cf. {\bf BBAF} and {\bf BBAT}).
{\bf Key:} {\em Fixed thresholds:} shown in boxplots of WHDR values for 50 simulated test sets for the two fixed
thresholds, and green bars are the value for the standard test set.  {\em Oracle thresholds:} heavy black bar.  {\em
  Held out threshold:} heavy red bar.  {\em Oracle threshold without smoothing:} heavy blue dashed bar.  {\em Fixed threshold
  without smoothing:} heavy purple bar.  {\em Boxplots:} horizontal bar = median; notch = fraction of interquartile
range outside which a difference in medians is significant; bottom and top of the box = 25 and 75 percentiles resp.;
whiskers extend to the most extreme data points that are not outliers; outliers -- greater than 1.5 times the
interquartile range outside top and bottom -- are '+'. 
{\em Best viewed in color.}
\label{smoothing}
}
\end{figure*}

\subsection{Simulated Test WHDR}

Because our method does not see any WHDR labels in training, we can estimate how WHDR reports change with test set.  We repeatedly
draw a simulated test set from the IIW dataset.  We then compute WHDR on each of the collection of simulated test sets
for each method.  This exposes the variance in WHDR caused by choice of test set.   For each simulated test set, each
image is chosen with probability $0.2$, yielding simulated test sets that are the same size as the standard test set.
We draw 50 simulated test sets to form a collection.  Results are shown as box plots of WHDR for all sets in the
collection at the two fixed thresholds.

{\bf Our methods are strong:} Figure~\ref{best} summarizes results for our strongest methods.  All beat rescaled Retinex.    The reported WHDR is
not particularly sensitive to threshold (note how held-out threshold WHDR is very close to oracle WHDR).  
There is some evidence that the standard test set is easier than randomly selected sets of the same
size (green bars in Figure~\ref{best} are mostly well below the median in the boxplot, and this is consistent across the figures).  

{\bf Adversarial smoothing is important:} Figure~\ref{smoothing} compares various configurations.  Adversarial smoothing is important to the method's success
({\bf NoSmo} is relatively weak, but better than both Retinex and \cite{Liu_2020_CVPR}).  

{\bf Averaging is very important:}  Both Figure~\ref{best} and Figure~\ref{smoothing} support the conclusion
that methods that use discrete image averaging and also average over more boxes and more scales
work noticeably better.  The notches on the boxplots allow judgements of significance; the difference between {\bf BBAF}
at $0.165$ and the other models is clearly significant.

\begin{figure}
  \includegraphics[width=\columnwidth]{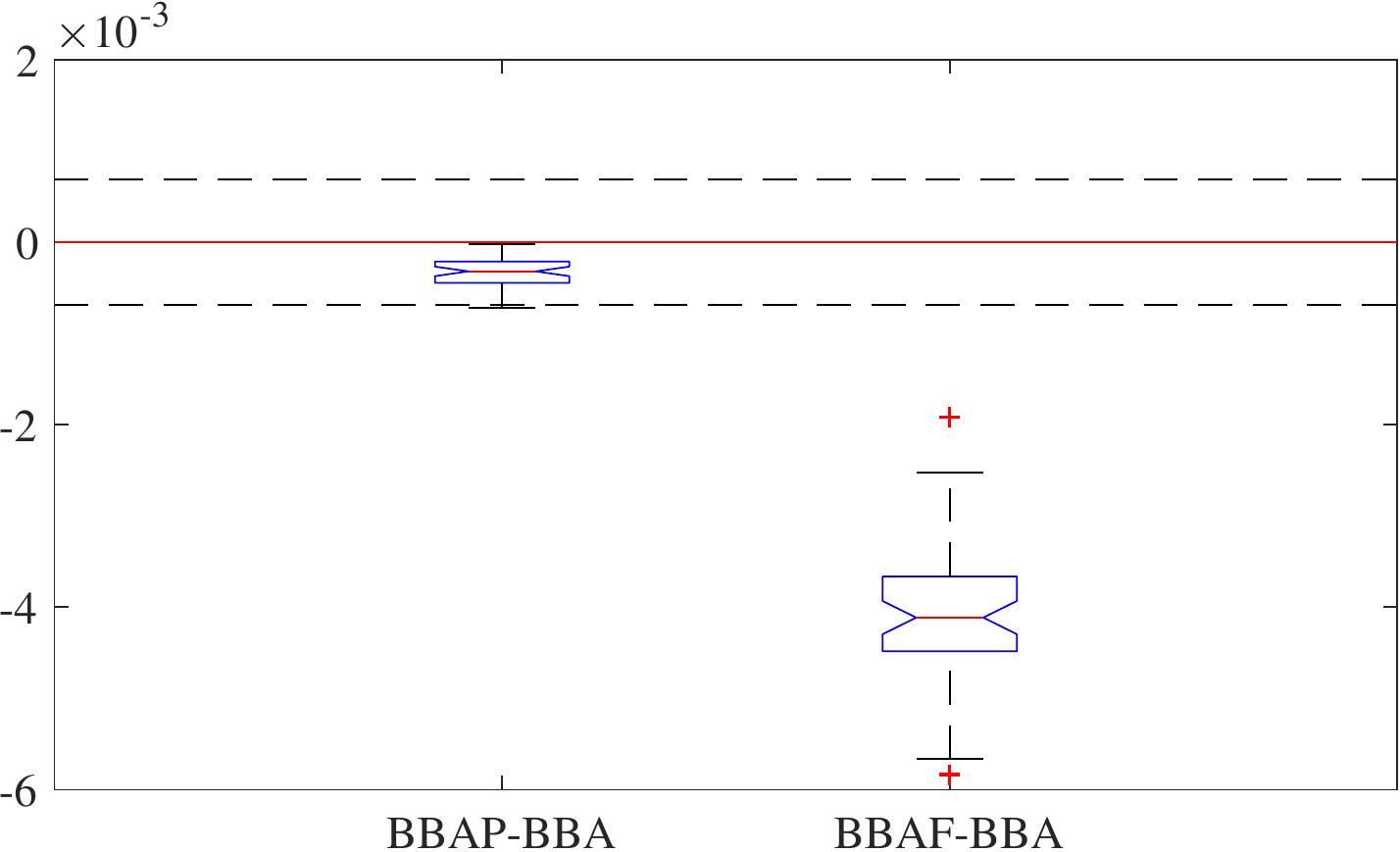}
\caption{\em Boxplots of the difference in WHDR for simulated test sets reported by pairs of models reveals 
whether one is reliably better than another; here {\bf BBAP}  almost always reports a slightly smaller WHDR than {\bf BBA},
and {\bf BBAF} always reports a very much better WHDR than {\bf BBA}.  
  The dashed lines show the three standard deviation range for the variation caused by random offsets in the averaging
  process.  The only question of significance is for the comparison between {\bf BBA} and {\bf BBAP}.  But random
  offsets in the averaging process should affect each method equivalently, and every difference favors {\bf
    BBAP}, suggesting that {\bf BBAP} is genuinely better than {\bf BBA}.   The difference between {\bf BBAF} and {\bf
    BBA} is pronounced.  {\bf BBAF} is clearly the best of our current models.
  {\em Best viewed in color.}
  \label{treat}
}
\end{figure}

{\bf Standard test set WHDR is unreliable:}  The WHDR varies quite strongly across simulated test sets -- the standard deviation is 0.3\% for the base method, but
note the quite heavy tails.  As a result, comparing methods using a single WHDR is unwise.  For example, the held-out
threshold WHDR for {\bf BBAP} appears to be worse than that for {\bf BBA} -- postprocessing appears to make things
worse, though qualitative differences strongly favor {\bf BBAP} (Figure~\ref{treat}).  Closer analysis reveals this is
misleading.

\begin{figure*}
\includegraphics[width=\textwidth]{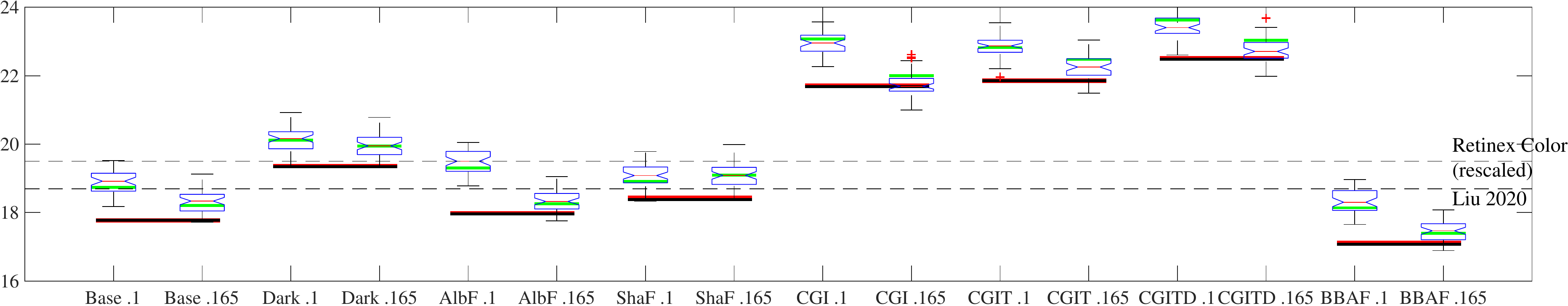}
\caption{\em  Varying the details of the paradigm has some effect; a {\bf Dark} shading paradigm creates notable difficulties, but varying the size of shading ({\bf ShaF}) and albedo ({\bf ShaF}) fragments seems to have only minor effects. Using tiles excerpted from CGIntrinsics \protect \cite{Li:2018if} leads to significant fall off in performance ({\bf CGI} -- tiles extracted from CGIntrinsics at original scale; {\bf CGIT} -- extracted from images shrunk so that tiles contain more detaile; {\bf CGITD} -- dependency between shading and albedo preserved).
  Graphical conventions as in Figure \protect \ref{best}.  
  {\em Best viewed in color.}
  \label{cgibad}
}
\end{figure*}

{\bf Paradigms are better than CGI:} Figure~\ref{cgibad} compares results for different paradigms and for CGI tiles.
The details of the paradigm do not seem to matter very much, but using CGIntrinsics for paradigm data causes a sharp loss in WHDR, resulting in methods that are outperformed by
Retinex.  Relative robustness to details of the paradigm is convenient, because we have no procedure to search
paradigms.  The paradigms described here are not the result of any systematic search.

\begin{figure*}
\includegraphics[width=\textwidth]{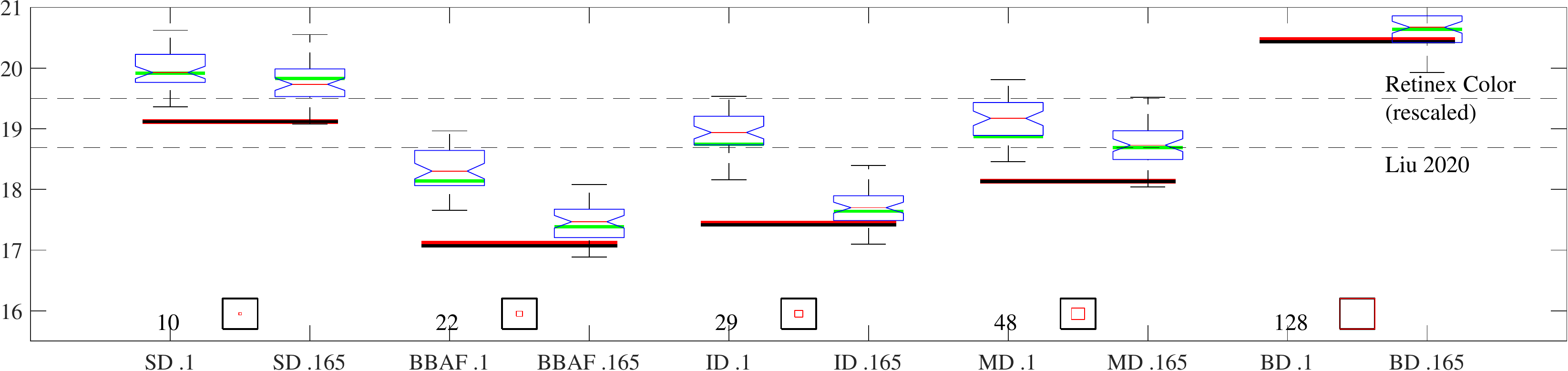}
\caption{\em  Varying the scale of the discriminator has an important effect on performance.  {\bf SD} the discriminator
  sees $10\times 10$ patches; {\bf BBAF} as in other figures our best model, $22\times 22$; {\bf ID} $29 \times 29$;
  {\bf MD} $48 \times 48$; and {\bf BD} $128\times 128$.  The scale of {\bf ID} was chosen by interpolating oracle WHDR
  for the others, then choosing the scale that produced the best predicted WHDR.  The red boxes show the scale of the
  discriminator patches with respect to the tile (black boxes) for each model.
  Graphical conventions as in Figure \protect \ref{best}.  
  {\em Best viewed in color.}
  \label{scalefx}
}
\end{figure*}

{\bf Scale is important:}  The discriminator is engineered to see albedo and shading patches of fixed size (the scale;
section~\ref{discscale}).  This parameter is important.   Figure~\ref{scalefx} shows performance for discriminators that view patches of several
different sizes.  The scale of discriminator patches has a strong effect on performance (Figure~\ref{scalefx}),  so that
imposing the requirement that a predicted albedo (resp. shading) looks like a paradigm at the wrong scale leads to a
notable fall-off in performance.   This -- and the tremendous improvements resulting from averaging -- supports the idea that
paradigms are essentially local models.

\begin{figure*}
\centerline{
\includegraphics[width=0.8\textwidth]{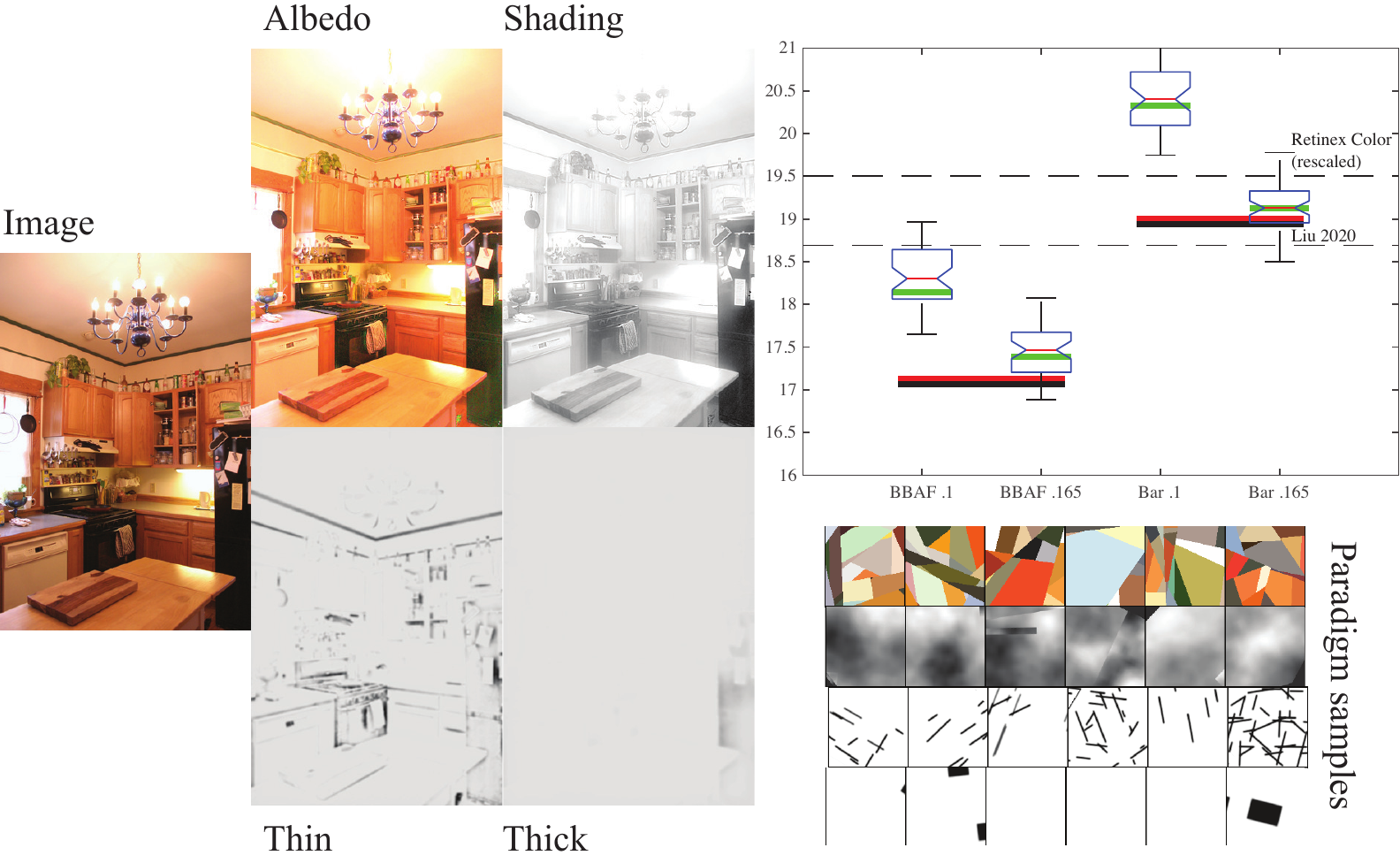}
}
\caption{\em  The method can be extended to capture thin and thick bars of darkness by extending the decomposer to have
  four heads (albedo, shading, thin bars, thick bars), and extending the paradigms ({\bf bottom left} shows examples).
  The advantage of doing so is that a decomposition will then capture the thin bars of darkness associated with grooves
  separately from albedo (example decomposition shown here).  Qualitatively, these thin bars do appear to be associated
  with grooves (but note the thin dark paint bars on the ceiling, which also appear in this map).  The cost in WHDR
  ({\bf top right} compares to {\bf BBAF}) is noticeable, but may be tolerable in some applications. 
  {\em Best viewed in color.}
  \label{bars}
}
\end{figure*}
{\bf Which model is best:} Note that standard test set WHDR suggests that {\bf BBA} is our best model
(Figure~\ref{best}; Table~\ref{numbers}).   Standard test set WHDR is a poor way to choose models, because WHDR varies quite strongly
across simulated test sets, and because the standard test set seems to be somewhat easy than randomly selected test
sets.  From Figure~\ref{best}, {\bf BBA}, {\bf BBAP} and {\bf BBAF} are reasonable contenders for best model.  
Figure~\ref{treat} shows a {\em treatment effects} comparison of these models.
For each simulated test set, we compute the difference between WHDR reported by the two models
($\mbox{W}_A-\mbox{W}_B$). If a boxplot of these differences straddles $0$, the models may be the same; if it lies far
below (resp. above) $0$, then model A (resp. B) is better, because on most simulated test sets it gets lower WHDR.
Figure~\ref{treat} shows these boxplots comparing {\bf BBA}, {\bf BBAP}, {\bf BBAF}.   {\bf BBAP} appears slightly
better than {\bf BBA}; {\bf BBAF} is clearly a lot better, because for every simulated test set, the WHDR of its
predictions is below that of {\bf BBA}.  Some variation in the reported difference in WHDR must be
caused by the random offsets in the averaging process of section~\ref{covariance}.  We estimate this variance by computing WHDR for different
averages of the same simulated test set.   The figure suggests that this effect is not strong enough to explain the
difference between {\bf BBAF} and {\bf BBA}.

\begin{figure}
\includegraphics[width=\columnwidth]{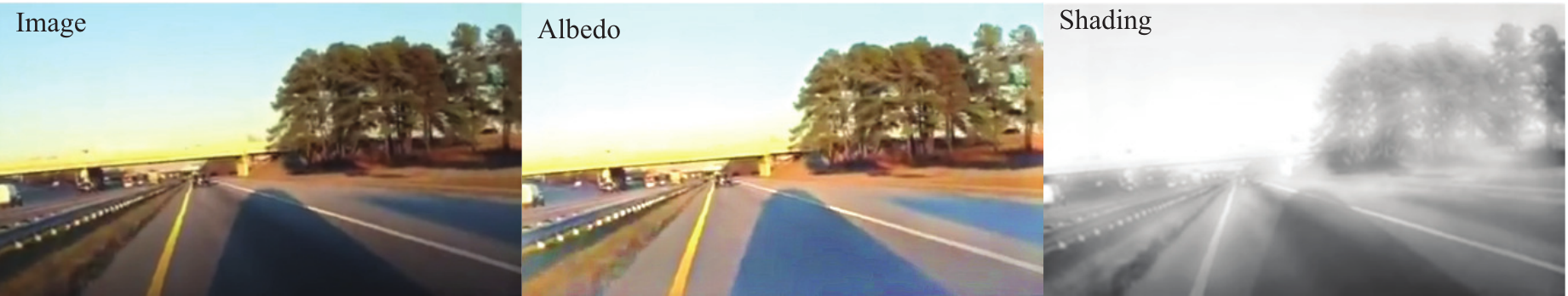}
\caption{\em  Our method suffers indecisiveness, as do others; this is a persistent problem in intrinsic image methods.
  Figures show a decomposition of an outdoor image, using our method.  Note the pronounced shadow leaves effects in both
  albedo and shading fields; versions of this effect for other methods can be seen in Figure \protect \ref{multiqual}.
  {\em Best viewed in color.}
  \label{mym}
}
\end{figure}

\subsection{Qualitative evaluation}

Figure~\ref{qualitative} shows some qualitative example albedos.  Note the method can suppress strong indoor shadows;
backscatter from shiny fittings; fast shading effects in clothing folds; and handle dark shadows well.  Figure~\ref{multiqual} shows comparisons to a number of strong
recent methods.  As these comparisons indicate, WHDR may be a limited guide to success.  Methods that achieve strong
WHDR on test can produce quite eccentric albedo fields.  One difficulty comes with choice of colors:  methods that do not enforce a small residual can produce
quite odd colors in the albedo field.  For this particular example, the method of \cite{Shi17} produces very strongly
saturated colors (and has a very poor WHDR).  The method of \cite{nestmeyer2017reflectance} (which gets a strong WHDR on
this scene) produces highly desaturated colors; \cite{EGSR18} have a better WHDR and somewhat less desaturated colors.
These methods uses postprocessing procedures to impose a piecewise constant albedo.  While this results in WHDR
improvements, the resulting albedo fields may be hard to use.  In particular, they display a
a ``colored paper'' effect, where surfaces look as though they are made of flat colored paper.
Figure~\ref{shengcompare} shows comparisons to other methods on a demanding outdoor image with dark shadows.  The method 
of~\cite{Sheng20} produces very strong shading recovery at the cost of a strong ``colored paper'' effect.

A piecewise constant albedo is entirely consistent with the spatial model underlying intrinsic image
estimation.   However, relatively few surfaces {\em behave} as if they have piecewise constant albedo.   
For example, the cupboard doors in Figure~\ref{multiqual} likely do not have piecewise constant albedo, and removing the
wood grain effect to the shading -- as the method of~\cite{EGSR18} does -- is likely a mistake.  The grain is the result
of real variations in albedo.  More difficult is handling shading at narrow grooves in surfaces (for example, between
the cabinet drawers in Figure~\ref{multiqual}).  The narrow dark shadows here are clearly a shading effect, but they
behave like an albedo effect.  As \cite{Koenderink} noted, deep grooves are hard to illuminate and so are dark for
almost all shading fields, an effect confirmed in~\cite{ICCVshade}.  This means they behave very largely like intrinsic
properties.  Whether these effects belong in an albedo map or a shading map likely depends on the application.  If, for
example, one wishes to report physical albedo, then they should appear in albedo effect,  though this application is uncommon.  
Alternatively, one may wish to reshade images, insert objects, and so on.  For these applications, it is likely better
for these effects to appear in albedo.  No current normal recovery method can resolve these effects, and leaving them
out of albedo means reshading will omit them. One alternative is to recover them in a separate layer, distinct from
shading and albedo.  Our method can do this, using a straightforward extension (Figure~\ref{bars}).
While quantitative evaluation methods for this kind of decomposition do not exist, qualitatively the thin bars appear in
sensible places, at some cost in WHDR.

As the qualitative comparison shows, all intrinsic image methods suffer from indecisiveness.  Albedo and shading reports
are quite strongly correlated, likely because nothing forces the method to ``make up its mind'' -- a shadow typically
results in a dark patch in both albedo and shading (for example, the dark fridge in Figure~\ref{multiqual}; Figure~\ref{mym}).  While this
does not appear to cause problems for WHDR score, reports with this property must be inaccurate.  Our method is somewhat
less subject to this effect than the others shown in Figure~\ref{multiqual} for that image.  However, the effect appears strongly
in Figure~\ref{shengcompare} for ours and other methods.

\pagebreak
\pagebreak

\section{Conclusion}

This paper has demonstrated a novel approach to intrinsic image decomposition.  The method relies entirely on authored
spatial models of the intrinsic components required.  These paradigms serve as a convenient encoding of priors.  A
decomposition network is trained to (a) network decompose authored paradigm images correctly and (b) produce albedo and shading
layers for real images that are ``like'' paradigms at short spatial scales.  Long scale error control is by a process of
averaging over translations, rotations and scales.  The method achieves better WHDR than any current unsupervised
method.  Qualitative evaluation suggests that the methods albedo maps may have advantages in computational photography
applications, as they do not display ``colored paper'' effects and they do capture groove shading as an intrinsic
(rather than extrinsic) phenomenon.   The method can be extended to represent other intrinsic effects, by supplying
spatial paradigms.


%

\appendices
\section{Network Details}

We have engaged in no organized search over network architectures, and do not claim either network to be
optimal. 

\subsection{Decomposer}

The decomposer is our implementation of a U-net with skip connections.  The encoder accepts $128\times 128 \times 7$
tiles (3 color dimensions, 4 from the location code).  A single layer of 1x1 convolutions increases the dimension of the
input, which is then subjected to five convolutional layers, each of kernel size 4, stride 2 and no padding.  Each layer
uses a leaky ReLU (0.2) as a nonlinearity; it is possible a ReLU would have been a better choice~\cite{ganrelu}, but we have
had no problems with stability in training.    The decoder accepts the resulting code, and applies five down
layers, each of kernel size 5, stride 1, and padding 2.  A down layer consists of: stacking the input
block with the corresponding block from the encoder (which has the same set of spatial dimensions, whence the choice of 
padding and stride), applying a convolutional layer to the result, then upsampling by 2 using a bilinear interpolate.
Finally, a 1x1 convolution projects to 3 dimensions, and a tanh nonlinearity is applied.

\subsection{Discriminator}

The standard discriminator consists of four convolutional layers, each with kernel size 4 and stride 2; there is no
padding, and there is a bias.  The nonlinearity is a leaky ReLU (0.2) and each layer is spectrally normalized.  This
produces an $8 \times 8 \times 1$ block of activations $u$. The training loss for the discriminator
is computed by averaging $\max \left(0, 1+y u\right)$ over this block, with $y=1$ when the batch 
consists of generated images and $y=-1$ when it consists of real images.  The loss that the 
discriminator produces for the generator is computed for a batch of generated images, and 
is computed by averaging $\max \left(0, 1- u\right)$ over this block,

\section{Weights and Codes}

{\bf Location codes:}  Each RGB tile is
stacked with four code tiles.   Each code tile represents distance to one of the four edges of the RGB tile, with the $(i, j)$'th 
location in the $k$'th code tile containing $\mbox{max}(0, 40-\mbox{dist}([i, j], \mbox{edge}_k)$.

{\bf Weighting window:}  The weight window is the pointwise minimum of four separate $x$ and $y$ weighting windows. 
The $(i, j)$'th pixel of the first $x$-weighting window is 
\[
\begin{array}{cc}
\frac{1-e^{41-j}}{1-e^{-1}}& j<=40\\
1& \mbox{otherwise}
\end{array}.
\]
The second $x$-weighting window is a reflection of the first; the $y$-weighting windows are transposes of the
$x$-weighting windows.



\section{Polishing Albedo and Shading}

Write $\vect{i}$ for the image at some pixel, $\vect{a}$ for the albedo estimate at that location (which is colored,
hence a vector), $s$ for the shading estimate.  We wish to compute updates to albedo and shading so that
\[
\vect{i}-(\vect{a}+\delta \vect{a})(s+\delta s)=0
\]
to first order.
Write $\vect{r}=\vect{i}-\vect{a}s$.  Then
\[
  \vect{r}=\vect{a}\delta s+ s \delta \vect{a}.
\]
This is underconstrained, so we seek to minimize $(\delta s)^2+(\delta \vect{a})^T(\delta \vect{a})$.  Substituting
$\delta \vect{a}=(1/s) (\vect{r}-\vect{a}\delta s)$ and minimizing yields the result.

\section{Losses}
\label{losses}

Write $\al$ (resp. $\sh$, $\co$) for true and $\alp$ (resp. $\shp$, $\cop$) for predicted albedo (resp. shading, color)
for image $\im$.
We have 
\begin{eqnarray*}
{\cal L}_T(\theta)&=&\alpha_{a} {\cal L}_{a}+\alpha_{s}{\cal L}_{s}+\alpha_c {\cal L}_c +\alpha_r {\cal L}_{r}\\
&=&\alpha_a \left\{\frac{1}{N_b} \sum_{i} C(\al, \alp)\right\}+\\
&&\alpha_s \left\{\frac{1}{N_b} \sum_{i} C(\sh, \shp)\right\}+\\
&&\alpha_c \left\{\frac{1}{N_b} \sum_i C(\co, \cop)\right\}+\\
&&\alpha_r \left\{\frac{1}{N_b} \sum_i C(\mbox{render}(\alp, \shp, \cop), \im)\right\}
\end{eqnarray*}
where $C$ compares two fields (we use the mixed $L_1$-$L_2$ loss of ~\cite{} for albedo and shading, and $L_2$ for
color).

For real images, true albedo (resp. shading) is not known.  We assume the illuminant for real images is not colored. 
Write $l(x, t)=\left[\mbox{min}(0, x-t)\right]^2$ 
and $L_{E}(\im, t)=\sum_{uv} l(\im_{uv}, t)$. We use
\begin{eqnarray*}
{\cal L}_R (\theta)&=&\alpha_{rr} {\cal L}_{rr}+{\cal L}_{\mbox{range}}\\
&=&\alpha_{rr} \left\{\frac{1}{N_b} \sum_{i} C(\mbox{render}(\alp, \shp, \cop), \im)\right\}+\\
&&\alpha_{rc} \left\{\frac{1}{N_b}\sum_j \left[C(\vect{c}(\ipt_i; \theta), \vect{1})\right]\right\}+\\
&&\left\{\frac{1}{N_b} \sum_i \left[ L_{E}(\alp, 1)+L_{E}(\shp, 1)\right]\right\}+\\
&&\left\{\frac{1}{N_b} \sum_i \left[ L_{E}(-\alp, 0)+L_{E}(-\shp, 0)\right]\right\}
\end{eqnarray*}

\section{Adversarial Smoothing}

Our procedure is as follows.  Write $\theta$ for the generator's parameters and $\phi$ for the discriminator's parameters.   
Write $\rb$ for a batch of $N$ real pairs, $\rtu_i$ for the $i$'th example from that batch, 
$\tb$ for a batch of training pairs, etc., $g(\cdot; \phi)$ for the discriminator (a parametric function that maps a pair to a
number) and $h(x, y)=max(0, 1-xy)$ for the hinge loss.  Note that $\rtu$ is a function of the map parameters $\theta$,
because it was generated by applying the map to an image.  Assume that we have estimates $\theta_k$, $\phi_l$ of
$\theta$, $\phi$.  We update $\phi_l$ by taking an optimizer step using the gradient
\[
\frac{1}{N} \sum_i \nabla_\phi \left(h(g(\ttu_i; \phi), -1)\right)
+\frac{1}{N} \sum_i \nabla_\phi \left(h(g(\rtu_i; \phi), 1)\right)
\]
to obtain $\phi_{l+1}$.  We now add the following term to the gradient with respect to $\theta$:
\[
\alpha_d \frac{1}{N} \sum_i \nabla_\theta \left(g(\rtu_i(\theta); \phi_{l+1})\right)
\]
and update $\theta_k$ by taking an optimizer step using the resulting gradient ($\alpha$ is as before).



\ifCLASSOPTIONcaptionsoff
  \newpage
\fi



%

\bibliographystyle{IEEEtran}
\bibliography{dafjulia,lana,dafsupp,bigdaf,bib_mani,references,references-mani,jasonrefs,main}

\begin{thebibliography}{10}
\providecommand{\url}[1]{#1}
\csname url@samestyle\endcsname
\providecommand{\newblock}{\relax}
\providecommand{\bibinfo}[2]{#2}
\providecommand{\BIBentrySTDinterwordspacing}{\spaceskip=0pt\relax}
\providecommand{\BIBentryALTinterwordstretchfactor}{4}
\providecommand{\BIBentryALTinterwordspacing}{\spaceskip=\fontdimen2\font plus
\BIBentryALTinterwordstretchfactor\fontdimen3\font minus
  \fontdimen4\font\relax}
\providecommand{\BIBforeignlanguage}[2]{{%
\expandafter\ifx\csname l@#1\endcsname\relax
\typeout{** WARNING: IEEEtran.bst: No hyphenation pattern has been}%
\typeout{** loaded for the language `#1'. Using the pattern for}%
\typeout{** the default language instead.}%
\else
\language=\csname l@#1\endcsname
\fi
#2}}
\providecommand{\BIBdecl}{\relax}
\BIBdecl

\bibitem{BarTen78}
H.~Barrow and J.~Tenenbaum, ``Recovering intrinsic scene characteristics from
  images,'' in \emph{Comp. Vision Systems}, 1978.

\bibitem{nn8146}
E.~Land, ``Color vision and the natural image: Part i,'' vol.~45, no.~1, pp.
  115--129, January 1959.

\bibitem{nn8147}
------, ``Color vision and the natural image: Part ii,'' vol.~45, no.~4, pp.
  636--644, April 1959.

\bibitem{mccann1976quantitative}
J.~J. McCann, S.~P. McKee, and T.~H. Taylor, ``Quantitative studies in retinex
  theory a comparison between theoretical predictions and observer responses to
  the “color mondrian” experiments,'' \emph{Vision research}, 1976.

\bibitem{horn1974determining}
B.~K. Horn, ``Determining lightness from an image,'' \emph{Computer graphics
  and image processing}, 1974.

\bibitem{horn1973lightness}
------, ``On lightness,'' 1973.

\bibitem{Blake}
A.~Blake, ``Boundary conditions for lightness computation in mondrian world,''
  \emph{Computer Vision, Graphics and Image Processing}, vol.~32, pp. 314--327,
  1985.

\bibitem{BrelstaffBlake}
G.~Brelstaff and A.~Blake, ``Computing lightness,'' \emph{Pattern Recognition
  Letters}, vol.~5, no.~2, pp. 129--38, 1987.

\bibitem{kimmel2003variational}
R.~Kimmel, M.~Elad, D.~Shaked, R.~Keshet, and I.~Sobel, ``A variational
  framework for retinex,'' \emph{International Journal of computer vision},
  vol.~52, no.~1, pp. 7--23, 2003.

\bibitem{elad03}
M.~Elad, R.~Kimmel, D.~Shaked, and R.~Keshet, ``Reduced complexity retinex
  algorithm via the variational approach,'' \emph{J Vis Commun Image R},
  vol.~14, no.~4, pp. 369--388, 2003.

\bibitem{weiss2001deriving}
Y.~Weiss, ``Deriving intrinsic images from image sequences,'' in
  \emph{Proceedings Eighth IEEE International Conference on Computer Vision.
  ICCV 2001}, 2001.

\bibitem{levin2004separating}
A.~Levin, A.~Zomet, and Y.~Weiss, ``Separating reflections from a single image
  using local features,'' in \emph{Proceedings of the 2004 IEEE Computer
  Society Conference on Computer Vision and Pattern Recognition, 2004. CVPR
  2004.}, 2004.

\bibitem{Levin:2007ho}
A.~Levin and Y.~Weiss, ``{User Assisted Separation of Reflections from a Single
  Image Using a Sparsity Prior},'' \emph{IEEE TPAMI}, vol.~29, no.~9, pp.
  1647--1654, Jan. 2007.

\bibitem{zhao2012closed}
Q.~Zhao, P.~Tan, Q.~Dai, L.~Shen, E.~Wu, and S.~Lin, ``{A closed-form solution
  to retinex with nonlocal texture constraints},'' \emph{IEEE TPAMI}, vol.~34,
  no.~7, pp. 1437--1444, 2012.

\bibitem{Tappen:2005hy}
M.~F. Tappen, W.~T. Freeman, and E.~H. Adelson, ``{Recovering intrinsic images
  from a single image},'' \emph{IEEE Transactions on Pattern Analysis and
  Machine Intelligence}, vol.~27, no.~9, pp. 1459--1472, 2005.

\bibitem{brainard1997bayesian}
D.~H. Brainard and W.~T. Freeman, ``Bayesian color constancy,'' \emph{JOSA A},
  1997.

\bibitem{cc42423}
M.~Farenzena and A.~Fusiello, ``Recovering intrinsic images using an
  illumination invariant image,'' in \emph{ICCV}, 2007, pp. III: 485--488.

\bibitem{brainard:1986an}
D.~Brainard and B.~Wandell, ``Analysis of the retinex theory of color vision,''
  \emph{J. Opt. Soc. America - A}, vol.~3, 1986.

\bibitem{Bousseau:2009ec}
A.~Bousseau, A.~Bousseau, S.~Paris, and F.~Durand, ``{User-assisted intrinsic
  images},'' in \emph{ACM Transactions on Graphics (TOG)}.\hskip 1em plus 0.5em
  minus 0.4em\relax ACM, Dec. 2009, p. 130.

\bibitem{Chang:2014ti}
J.~Chang, R.~Cabezas, and J.~W. Fisher~III, ``Bayesian nonparametric intrinsic
  image decomposition,'' \emph{Proceedings of the European Conference on
  Computer Vision (ECCV)}, Sept 2014.

\bibitem{Gehler:2011vc}
P.~V. Gehler, C.~Rother, M.~Kiefel, L.~Zhang, and B.~Scholkopf, ``{Recovering
  Intrinsic Images with a Global Sparsity Prior on Reflectance.}'' in
  \emph{Advances in neural information processing systems}, 2011, pp. 765--773.

\bibitem{shen2011intrinsic}
J.~Shen, X.~Yang, Y.~Jia, and X.~Li, ``{Intrinsic images using optimization},''
  in \emph{Computer Vision and Pattern Recognition}.\hskip 1em plus 0.5em minus
  0.4em\relax IEEE, 2011, pp. 3481--3487.

\bibitem{Shen:2011gf}
L.~Shen and C.~Yeo, ``{Intrinsic images decomposition using a local and global
  sparse representation of reflectance.}'' \emph{CVPR}, 2011.

\bibitem{Barron:2013uc}
J.~T. Barron and J.~Malik, ``Shape, illumination, and reflectance from
  shading,'' EECS, UC Berkeley, Tech. Rep. UCB/EECS-2013-117, May 2013.

\bibitem{Sheng20}
B.~Sheng, P.~Li, Y.~Jin, P.~Tan, , and T.-Y. Lee, ``Intrinsic image
  decomposition with step and drift shading separation,'' \emph{TPAMI}, 2020.

\bibitem{Shi17}
J.~Shi, Y.~Dong, H.~Su, and S.~X. Yu, ``Learning nonlambertian object
  intrinsics across shapenet categories,'' in \emph{CVPR}, 2017.

\bibitem{EGSR18}
S.~Bi, N.~K. Kalantari, and R.~Ramamoorthi, ``Deep hybrid real and synthetic
  training for intrinsic decomposition,'' in \emph{Eurographics Symposium on
  Rendering}, 2018.

\bibitem{Zhou:2015fp}
T.~Zhou, P.~Krahenb{\"u}hl, and A.~A. Efros, ``{Learning data-driven
  reflectance priors for intrinsic image decomposition},'' in
  \emph{International Conference on Computer Vision}.\hskip 1em plus 0.5em
  minus 0.4em\relax IEEE, 2015, pp. 3469--3477.

\bibitem{narihira2015regress}
T.~Narihira, M.~Maire, and S.~X. Yu, ``Learning lightness from human judgement
  on relative reflectance,'' in \emph{Proceedings of the IEEE CVPR}, 2015.

\bibitem{Zhou_2015_ICCV}
T.~Zhou, P.~Krahenbuhl, and A.~A. Efros, ``Learning data-driven reflectance
  priors for intrinsic image decomposition,'' in \emph{Proceedings of the IEEE
  International Conference on Computer Vision (ICCV)}, December 2015.

\bibitem{li2018cgintrinsics}
Z.~Li and N.~Snavely, ``Cgintrinsics: Better intrinsic image decomposition
  through physically-based rendering,'' in \emph{Proceedings of the European
  Conference on Computer Vision (ECCV)}, 2018, pp. 371--387.

\bibitem{8579030}
Q.~{Fan}, J.~{Yang}, G.~{Hua}, B.~{Chen}, and D.~{Wipf}, ``Revisiting deep
  intrinsic image decompositions,'' in \emph{CVPR}, 2018.

\bibitem{yu2019inverserendernet}
Y.~Yu and W.~A. Smith, ``Inverserendernet: Learning single image inverse
  rendering,'' in \emph{Proceedings of the IEEE Conference on Computer Vision
  and Pattern Recognition}, 2019.

\bibitem{bell14intrinsic}
S.~Bell, K.~Bala, and N.~Snavely, ``Intrinsic images in the wild,'' \emph{ACM
  Trans. on Graphics (SIGGRAPH)}, 2014.

\bibitem{Liu_2020_CVPR}
Y.~Liu, Y.~Li, S.~You, and F.~Lu, ``Unsupervised learning for intrinsic image
  decomposition from a single image,'' in \emph{Proceedings of the IEEE/CVF
  Conference on Computer Vision and Pattern Recognition (CVPR)}, June 2020.

\bibitem{Bi:2015eh}
S.~Bi, X.~Han, and Y.~Yu, ``{An L1 image transform for edge-preserving
  smoothing and scene-level intrinsic decomposition},'' \emph{ACM Transactions
  on Graphics}, vol.~34, no.~4, pp. 78--78:12, Jul. 2015.

\bibitem{Grosse:2009ji}
R.~Grosse, M.~K. Johnson, E.~H. Adelson, and W.~T. Freeman, ``Ground-truth
  dataset and baseline evaluations for intrinsic image algorithms,'' in
  \emph{International Conference on Computer Vision}, 2009, pp. 2335--2342.

\bibitem{Tappen:2006vo}
M.~F. Tappen, E.~H. Adelson, and W.~T. Freeman, ``{Estimating Intrinsic
  Component Images using Non-Linear Regression.}'' in \emph{Computer Vision and
  Pattern Recognition}.\hskip 1em plus 0.5em minus 0.4em\relax IEEE, 2006, pp.
  1992--1999.

\bibitem{barron2014shape}
J.~T. Barron and J.~Malik, ``Shape, illumination, and reflectance from
  shading,'' \emph{IEEE transactions on pattern analysis and machine
  intelligence}, 2014.

\bibitem{Sintel}
D.~J. Butler, J.~Wulff, G.~B. Stanley, and M.~J. Black, ``A naturalistic open
  source movie for optical flow evaluation,'' in \emph{ECCV}, 2012.

\bibitem{ChenKoltun}
Q.~Chen and V.~Koltun, ``A simple model for intrinsic image decomposition with
  depth cues,'' in \emph{ICCV}, 2013.

\bibitem{Lettry}
\BIBentryALTinterwordspacing
L.~Lettry, K.~Vanhoey, and L.~V. Gool, ``Deep unsupervised intrinsic image
  decomposition by siamese training,'' \emph{CoRR}, vol. abs/1803.00805, 2018.
  [Online]. Available: \url{http://arxiv.org/abs/1803.00805}
\BIBentrySTDinterwordspacing

\bibitem{}
M.~Dunham, \emph{Data Mining: Introductory and Advanced Topics}.\hskip 1em plus
  0.5em minus 0.4em\relax Prentice Hall, 2003.

\bibitem{bi20151}
S.~Bi, X.~Han, and Y.~Yu, ``An l 1 image transform for edge-preserving
  smoothing and scene-level intrinsic decomposition,'' \emph{ACM Transactions
  on Graphics (TOG)}, 2015.

\bibitem{nestmeyer2017reflectance}
T.~Nestmeyer and P.~V. Gehler, ``Reflectance adaptive filtering improves
  intrinsic image estimation,'' in \emph{Proceedings of the IEEE Conference on
  Computer Vision and Pattern Recognition}, 2017, pp. 6789--6798.

\bibitem{bilat}
J.~T. Barron and B.~Poole, ``The fast bilateral solver,'' in \emph{ECCV}, 2016.

\bibitem{Li:2018if}
Z.~Li and N.~Snavely, ``{CGIntrinsics - Better Intrinsic Image Decomposition
  Through Physically-Based Rendering.}'' \emph{ECCV}, vol. 11207, no.~4, pp.
  381--399, 2018.

\bibitem{narihira2015direct}
T.~Narihira, M.~Maire, and S.~X. Yu, ``Direct intrinsics: Learning
  albedo-shading decomposition by convolutional regression,'' in
  \emph{Proceedings of the IEEE international conference on computer vision},
  2015.

\bibitem{Janner}
M.~Janner, J.~Wu, T.~D. Kulkarni, I.~Yildirim, and J.~B. Tenenbaum,
  ``Self-supervised intrinsic image decomposition,'' in \emph{NIPS}, 2017.

\bibitem{Cheng_2018_CVPR}
L.~Cheng, C.~Zhang, and Z.~Liao, ``Intrinsic image transformation via scale
  space decomposition,'' in \emph{Proceedings of the IEEE Conference on
  Computer Vision and Pattern Recognition (CVPR)}, June 2018.

\bibitem{Weissseq}
Y.~Weiss, ``Deriving intrinsic images from image sequences,'' in \emph{ICCV},
  2001.

\bibitem{BigTimeLi18}
Z.~Li and N.~Snavely, ``Learning intrinsic image decomposition from watching
  the world,'' in \emph{Computer Vision and Pattern Recognition (CVPR)}, 2018.

\bibitem{Laffont_2015_ICCV}
P.-Y. Laffont and J.-C. Bazin, ``Intrinsic decomposition of image sequences
  from local temporal variations,'' in \emph{Proceedings of the IEEE
  International Conference on Computer Vision (ICCV)}, December 2015.

\bibitem{ma2018single}
W.-C. Ma, H.~Chu, B.~Zhou, R.~Urtasun, and A.~Torralba, ``Single image
  intrinsic decomposition without a single intrinsic image,'' in
  \emph{Proceedings of the European Conference on Computer Vision (ECCV)},
  2018.

\bibitem{augsurvey}
C.~Shorten and T.~Khoshgoftaar, ``A survey on image data augmentation for deep
  learning,'' \emph{J. Big Data}, vol.~6, no.~60, 2019.

\bibitem{qi1}
T.O.Binford, ``Inferring surfaces from images,'' \emph{Artificial Intelligence
  Journal}, 1981.

\bibitem{qi2}
------, ``Generic observability model,'' in \emph{Proc. lnt Symp on Robotics
  Research}, 1987.

\bibitem{CohenWelling}
T.~S. Cohen and M.~Welling, ``Group equivariant convolutional networks,'' in
  \emph{ICML}, 2016.

\bibitem{LencVedaldi}
K.~Lenc and A.~Vedaldi, ``Understanding image representations by measuring
  their equivariance and equivalence,'' \emph{Int J Comput Vis}, vol. 127, no.
  456–476, 2019.

\bibitem{GoodfellowARXIV2014}
I.~J. Goodfellow, J.~Pouget-Abadie, M.~Mirza, B.~Xu, D.~Warde-Farley, S.~Ozair,
  A.~Courville, and Y.~Bengio, ``{Generative Adversarial Networks},'' in
  \emph{https://arxiv.org/abs/1406.2661}, 2014.

\bibitem{pmlr-v70-arora17a}
S.~Arora, R.~Ge, Y.~Liang, T.~Ma, and Y.~Zhang, ``Generalization and
  equilibrium in generative adversarial nets ({GAN}s),'' in \emph{ICML}, 2017.

\bibitem{zhou2015learning}
T.~Zhou, P.~Krahenbuhl, and A.~A. Efros, ``Learning data-driven reflectance
  priors for intrinsic image decomposition,'' in \emph{Proceedings of the IEEE
  International Conference on Computer Vision}, 2015, pp. 3469--3477.

\bibitem{Koenderink}
J.~Koenderink and A.~V. Doorn, ``Geometrical modes as a method to treat diffuse
  interreflections in radiometry,'' \emph{J. Opt. Soc. Am.}, vol.~73, no.~6,
  pp. 843--850, 1983.

\bibitem{ICCVshade}
J.~Haddon and D.~Forsyth, ``Shading primitives,'' in \emph{ICCV}, 1997.

\bibitem{ganrelu}
A.~Radford, L.~Metz, and S.~Chintala, ``Unsupervised representation learning
  with deep convolutional generative adversarial networks,'' in \emph{ICLR},
  2016.

\end{thebibliography}

%








\end{document}